\documentclass{article}

\PassOptionsToPackage{numbers, compress}{natbib}

\usepackage{hyperref}
\usepackage[dvipsnames]{xcolor}
\definecolor{darkpastelgreen}{rgb}{0.01, 0.75, 0.24}
\hypersetup{
    colorlinks=true,
    citecolor=darkpastelgreen,
    linkcolor=red,
    urlcolor=blue
}

 \usepackage[numbers]{natbib}

\usepackage{wrapfig}
\usepackage{amsmath} 
\usepackage{graphicx}
\usepackage{caption}
\usepackage{subcaption}
\usepackage{adjustbox}
\usepackage{makecell} 
\usepackage{xltabular} 

 \usepackage[main, final]{neurips_2026}

\definecolor{partconcept}{HTML}{A64D79}
\definecolor{partconcept_encoder}{HTML}{38761D}

\newcommand{\HR}[1]{{\color{Bittersweet} \textbf{HR:} #1}}
\newcommand{\VA}[1]{{\color{ForestGreen} \textbf{VA:} #1}}
\newcommand{\RK}[1]{{\color{BlueViolet} \textbf{RK:} #1}}
\newcommand{\VK}[1]{{\color{Maroon} \textbf{VK:} #1}}
\newcommand{\TODO}[1]{{\color{BurntOrange} \textbf{TODO:} #1}}

\renewcommand{\HR}[1]{}
\renewcommand{\VA}[1]{}
\renewcommand{\RK}[1]{}
\renewcommand{\VK}[1]{}
\renewcommand{\TODO}[1]{}

\usepackage[utf8]{inputenc} 
\usepackage[T1]{fontenc}    
\usepackage{hyperref}       
\usepackage{url}            
\usepackage{booktabs}       
\usepackage{amsfonts}       
\usepackage{nicefrac}       
\usepackage{microtype}      
\usepackage{xcolor}         
\usepackage{multirow}
\usepackage{longtable} 
\usepackage{cleveref}

    \title{\textsc{PartConcepts}: A Unified Mechanism for Fine-Grained Part Localization and Generation}

\author{%
  Vaibhav Agrawal \\
  IIIT Hyderabad \\
  \And
  Varghese Kuruvilla \\
  IIIT Hyderabad \\
  \AND
  Harsh Rangwani \\
  Adobe Research, Bengaluru \\
  \And
  Ravi Kiran Sarvadevabhatla \\
  IIIT Hyderabad \\
}

\begin{document}

\maketitle
\vspace{-10mm}

\begin{figure}[h!] 
  \centering
  \includegraphics[width=\textwidth]{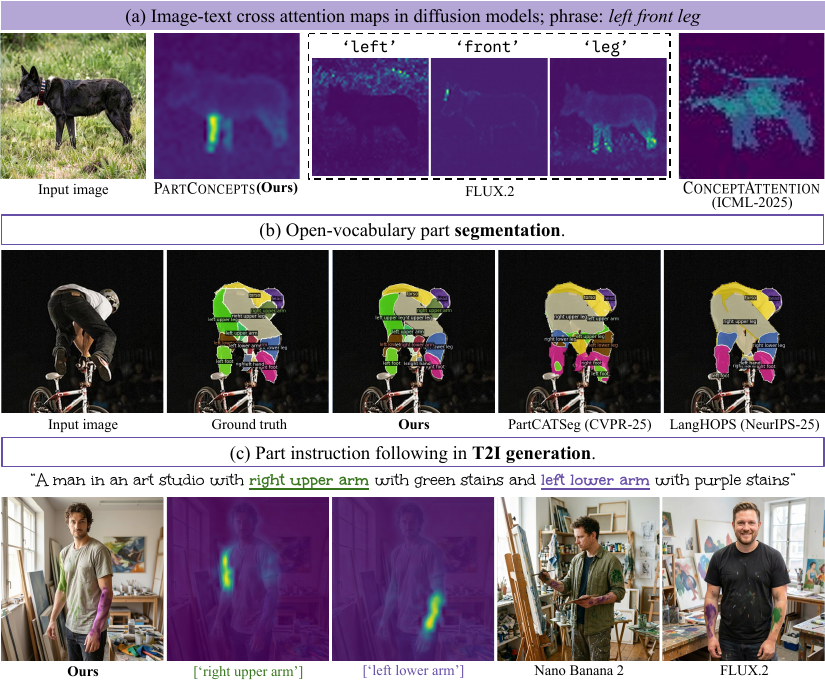} 
  \caption{\textbf{\textsc{PartConcepts}.} Cross attention maps of state-of-the-art diffusion models~\cite{flux-2-2025} reveal that they lack fine-grained part understanding, shown in (a). To address this, we propose \textsc{PartConcepts}, a unified mechanism for fine-grained part localization (see b.) and generation (see c.).} 
  \label{fig:teaser} 
\end{figure}

\begin{abstract}
While text-to-image (T2I) diffusion models exhibit strong semantic disentanglement at the object level, they struggle to localize fine-grained object parts. Although the latent visual features in these models are sufficiently fine-grained, the text–image interaction (cross-attention) fails to effectively exploit this information. This results in coarse and ambiguous localization, particularly for spatially distinct components such as `left' and `right' limbs. To address this limitation, we introduce \textbf{\textsc{PartConcepts}}, a mechanism that encodes textual part descriptions into learnable \textsc{PartConcept} tokens. These tokens are trained to selectively attend to their corresponding part regions in the image. To evaluate the part-level understanding of \textsc{PartConcept} tokens, we first probe its effectiveness for \textit{part segmentation}. We then evaluate whether the improved localisation enables \textit{fine-grained instruction following} in T2I generation. On standard Open-Vocabulary Part Segmentation (OVPS) benchmark, our method outperforms dedicated segmentation baselines, establishing a new state-of-the-art. We further evaluate our method on the more complex Pascal-Part benchmark for \textit{part instance segmentation} which contains fine-grained spatial annotations (e.g. `left lower arm'), outperforming all baselines significantly (\textbf{+15 mIoU}). Finally, our qualitative and quantitative results show that this strong localization of \textsc{PartConcept} tokens directly enables \textit{fine-grained instruction following} in T2I generation: the textual attributes (e.g., color) \textit{inherently} bind well to corresponding \textsc{PartConcept} tokens, without explicit attribute control mechanisms. This indicates that \textsc{PartConcept} tokens compose seamlessly with other text tokens, highlighting their applicability for fine-grained text-based generative control. Code and data are available on our \href{https://visionpartconcepts.github.io}{project page}.
\end{abstract}

\section{Introduction}

Text-to-image (T2I) diffusion models possess strong semantic priors that enable precise generation and editing~\cite{hertz2022prompt,avrahami2025stable}. Beyond generation, recent work~\cite{helbling2025conceptattention} shows that multi-modal Diffusion Transformer (mmDiT)~\cite{esser2024scaling,peebles2023scalable} cross-attention maps can serve as competitive object-instance segmentors. These results indicate that the diffusion latent space achieves effective semantic disentanglement at the object level. However, since objects are fundamentally defined by their constituent parts, any system aiming for part-level fine-grained reasoning and precise generative control must rely on semantic part-level representations. This raises a critical question: Does the image-text alignment of diffusion models actually extend to the part-level? Parts are uniquely challenging because quite often a single object contains multiple instances of the same component (e.g., `left ear' vs. `right ear'). Distinguishing these requires an understanding of object-centric spatial organization: the relative positioning of parts within the object's own frame of reference~\cite{zhang2023telling,mariotti2024improving}.

An investigation of the image-text attention maps for state-of-the-art models like FLUX.2~\cite{flux-2-2025} and \textsc{ConceptAttention}~\cite{helbling2025conceptattention} reveals that they are coarse and poorly localized for fine-grained part descriptions (see~\cref{fig:teaser}(a)). This poor localization manifests as a failure in downstream instruction following. As shown in~\cref{fig:teaser}(c), even state-of-the-art models fail to follow prompts targeting fine-grained parts. \textbf{Hence, we conclude that the image-text alignment in diffusion models, while strong at the object level, does not persist at the part level}. To identify the cause of failure, we analyze FLUX.2 and make two key observations:

\textbf{Observation \textsc{o1}.} The latent visual features of diffusion models are semantically very fine-grained, as evidenced by their ability to support highly accurate fine-grained image-to-image pointwise semantic correspondence tasks~\cite{zhang2023tale,tang2023emergent,zhang2023telling,hyperfeatures,hedlin2023unsupervised,gan2025unleashing}, with minimal training supervision. 

\textbf{Observation \textsc{o2}.} The text-image interaction (cross-attention) acts as a bottleneck; it fails to leverage the high-resolution latent visual features, resulting in coarse attention maps that cannot distinguish between spatially distinct parts like the `left' and `right' limbs, see~\cref{fig:teaser}(a).

These observations imply that while the latent features possess sufficient resolution (\textsc{o1}), the text-image interaction acts as a bottleneck (\textsc{o2}). Therefore, our core research question is: \textbf{can we improve text aligned part understanding of a T2I model, while preserving its strong priors?} To address this, we introduce \textsc{PartConcepts}. Given a text prompt, we encode each textual part description (e.g., `right upper arm') into a single \textsc{PartConcept} token embedding: this results in a structured token sequence, e.g., `A photo of an artist with \textsc{PartConcept}[`right upper arm'] \dots', which is passed to the mmDiT. During training, this \textsc{PartConcept} token is optimized to attend to the corresponding part region in the image, providing the fine-grained grounding that base models lack. As we show in our results, our proposed method effectively generalizes to novel part phrases, object classes and prompts beyond what is seen in the training data.


To evaluate the part-level understanding of \textsc{PartConcept} tokens, we first demonstrate our method's effectiveness for \textit{part segmentation}. We then show that this strong localization directly enables precise part-level attribute binding, thus providing \textit{fine-grained instruction following} in T2I generation. Thus, our \textsc{PartConcept}s provide a unified mechanism for both visual grounding and generative control. 

In the setting of \textit{part segmentation}, we extract the part segmentation maps directly from \textsc{PartConcept} attention maps, using only a single forward pass through the mmDiT network; this enables efficient inference without the need for iterative sampling. \textsc{PartConcepts} outperforms dedicated segmentation baselines on the standard Open-Vocabulary Part Segmentation (OVPS)~\cite{ov-parts} benchmark, establishing state-of-the-art. However, we find that OVPS is too coarse for precise grounding: it typically removes spatial descriptors and collapses distinct instances such as `left hand' and `right hand' into a single `hand' instance. Therefore, we evaluate for \textit{part instance segmentation} using the original Pascal-Part~\cite{pascalpart} benchmark, which retains these fine-grained part instances and corresponding spatial descriptions. On this more rigorous benchmark, our method significantly outperforms all baselines and establishes a new state-of-the-art. 

Finally, we utilize the effective localization of \textsc{PartConcept}s to enable \textit{fine-grained instruction following in text-to-image generation}. We find that our localization-focused training inherently enables the model to bind attributes to specific parts, without a specialized architectural mechanism for attribute control. For example, when provided with a prompt such as `an artist with \textsc{PartConcept}[`right upper arm'] colored green \dots' (see~\cref{fig:teaser}(c)), the model successfully binds the color attribute to the correct part region during the generative process. These results demonstrate that \textsc{PartConcept}s preserve the compositional nature of text space, while maintaining the high-fidelity priors of the base T2I generative model. Our model enables fine-grained instruction following to part descriptions not seen during training, indicating its strong generalization and prior preservation. 

\section{Related work}

\begin{figure}[t] 
  \centering
  \includegraphics[width=0.9198\textwidth]{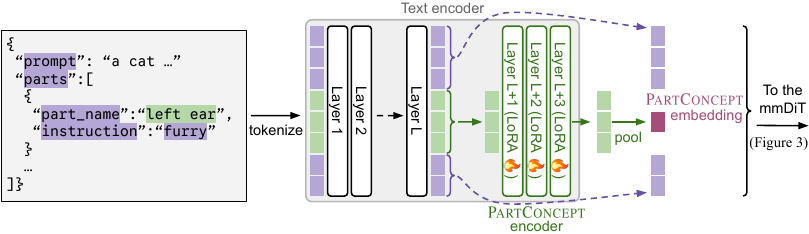} 
  \caption{\textbf{Encoding textual part descriptions into \textsc{PartConcept}s.} Given an input text prompt (left) with part phrases, our method encodes these textual part phrases (e.g. `left ear') into \textcolor{partconcept}{\textsc{PartConcept} embeddings}, which are trained to attend to corresponding image regions. We extend the text encoding in the base T2I model to encode the \textsc{PartConcept}s: the base T2I model uses intermediate text encoder outputs (at layer L) for conditioning the mmDiT; we use three subsequent text-encoder layers (L+1, L+2, L+3) for forming our \textcolor{partconcept_encoder}{\textsc{PartConcept} encoder}. We only train a LoRA on these subsequent layers; this design ensures that the output \textsc{PartConcept} does not drift too much from the text embedding space, thus preserving the T2I generative capabilities (see~\cref{subsec:experiments_t2i}).} 
  \label{fig:method_summarizer} 
  \vspace{-3mm} 
\end{figure}

\textbf{Semantic concepts in text-to-image (T2I) diffusion models.} Numerous studies have demonstrated that text-to-image (T2I) diffusion models encode strong semantic priors. These priors have also been shown to be effective for downstream tasks such as semantic correspondence~\cite{zhang2023tale,tang2023emergent,dunkel2025yourself}. A line of research works introduce learnable prompt tokens, and optimize them to attend to specific visual concepts in the image~\cite{avrahami2023break,liu2025partcomposer,unsupervisedkeypoints,hedlin2023unsupervised}. However, these works require training a separate embedding for every new visual concept, hence not scalable. Further, the learnt prompt tokens are typically not very composable with the text embeddings, hence reducing their applicability for localized attribute binding in T2I generation. A more scalable approach is to use the open vocabulary priors of the T2I models for learning visual concepts. Recent work, \textsc{ConceptAttention}~\cite{helbling2025conceptattention} proposes a training free framework for interpreting text-to-image multi-modal diffusion transformers (mmDiT), by extracting similarity maps of diffusion features with \textit{textual concept embeddings}. While these maps demonstrate strong object-level semantics, they lack fine-grained part-level understanding (see~\cref{fig:teaser}(a)), consistent with our observations. We attribute this limitation primarily to the weak alignment between textual concepts and visual features in the text-image interface of these models,  and propose an attention-based supervision strategy to mitigate this. 

\textbf{Part instance segmentation in images} is a long standing problem in computer vision~\cite{lazebnik2004semi,maji2013fine,vedaldi2014understanding,ade20ksceneparsing,partpanoptic,li2024panoptic,singh2022float}, with various datasets proposed for this task~\cite{pascalpart,he2022partimagenet,ramanathan2023paco,li2024partimagenet++}. Among these, Pascal-Part~\cite{pascalpart} stands out for its fine-grained part descriptions (e.g. `left ear', `right front upper leg'). Open-Vocabulary Part Segmentation (OVPS)~\cite{ov-parts} is the standard benchmark used by recent methods~\cite{pan2023towards,li2024partglee,wang2023hierarchical,choi2024understanding,choi2025fine,miao2025langhops}; however, OVPS is too coarse, as it removes spatial descriptors and collapses distinct instances such as `left hand' and `right hand' into a single `hand' instance (see Appendix~\ref{sec:ovps_vs_pascalpart}). Among recent works, LangHOPS~\cite{miao2025langhops} parses object hierarchies using an MLLM to generate queries which are decoded into part-level segmentation masks. PartCATSeg~\cite{choi2025fine} introduces a cost aggregation framework with a compositional loss and structural guidance using DINO~\cite{dinoemerging}. In contrast, our work leverages strong priors of generative models using a simple architecture to achieve fine-grained open-set part segmentation.

\textbf{Part level instruction following in T2I generation.} Part level instruction following has been explored in prior work~\cite{composingparts}, which uses self attention clustering followed by attribute injection through cross attention. However, it assumes image-text cross attention to be highly localized, which is violated in many cases, especially for fine-grained part descriptions. While there are works which focus on learning part-level concepts from images, and compose them for creative generations~\cite{liu2025partcomposer,ng2024partcraft}, they do not support text-based instruction following, hence not directly relevant. Recent work~\cite{cvejic2025partedit} uses learnable part concepts for localized image editing. However, it requires an embedding optimization procedure for every new part concept, hence not scalable. Further, it can only enable editing in a provided reference image. In contrast, our method directly enables instruction following in text-to-image generation, which is challenging due to absence of a reference structure.  

\section{Method}

\label{sec:method}

\begin{wrapfigure}{r}{0.475\textwidth}
  \centering
  \vspace{-10mm}
  \includegraphics[width=\linewidth]{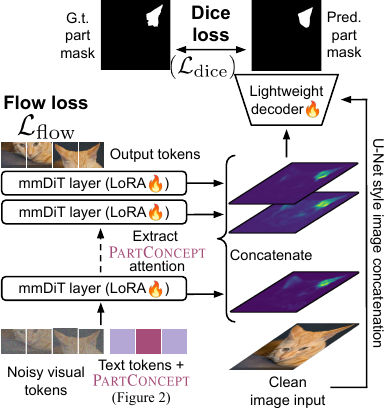} 
  \caption{\textbf{The localization objective.} We pass the interleaved text and \textcolor{partconcept}{\textsc{PartConcept}} tokens see~\cref{fig:method_summarizer}) through the text branch of the mmDiT model. During forward pass through the mmDiT, we extract layer-wise cross attention maps of the \textcolor{partconcept}{\textsc{PartConcept}} tokens. These attention maps, along with clean input image are decoded into precise part segmentation maps using a multi-scale CNN decoder, trained using dice loss. \textbf{The generative objective.} To preserve the generative capability of the T2I model, we also include the flow loss (denoising MSE), enabling fine-grained \textit{part-level instruction following} in T2I generation.} 
  \label{fig:method_training}
  \vspace{-10mm} 
\end{wrapfigure}

Given an input text prompt containing part phrases (e.g., `left ear', \cref{fig:method_summarizer}), our goal is to ensure that these phrases attend strongly to their corresponding image regions, thereby enabling both part localization and fine-grained instruction following in T2I generation.

In standard FLUX.2 cross-attention maps (\cref{fig:teaser}a), multi-word part phrases are split into discrete tokens that attend diffusely across disparate regions. Spatial descriptors (e.g., `left', `right', etc.) are especially dispersed, preventing the overall part description from localizing to a coherent visual region.


To address this, we encode the text embeddings of each part phrase into a single \textcolor{partconcept}{\textsc{PartConcept} embedding} of matching dimensionality (\cref{fig:method_summarizer}). Replacing discrete part-phrase tokens with their corresponding \textsc{PartConcept} token yields a structured sequence, which is fed into the multi-modal diffusion transformer (mmDiT) alongside noisy visual tokens (\cref{fig:method_training}). To prevent \textsc{PartConcept} embeddings from drifting away from the text manifold, we initialize the \textcolor{partconcept_encoder}{\textsc{PartConcept} encoder} using successive layers of the base text encoder and optimize only a LoRA~\citep{hu2022lora} adapter. As shown in~\cref{subsec:ablations}, this parameter-efficient design preserves the base T2I model's generative prior.

\textbf{The localization objective.} To localize \textsc{PartConcept} tokens to correct image regions, we optimize their cross-attention maps using a Dice segmentation loss~\citep{milletari2016v}. While \textsc{PartConcept} cross attention maps provide strong semantic localization, directly converting them into dense segmentation masks faces fundamental limitations: First, diffusion models operate over noisy latents, meaning the resulting cross attention maps inherently lack access to uncorrupted, fine-grained image evidence. Second, VAE latent compression restricts attention maps to a coarse spatial grid, precluding pixel-accurate contour prediction. Finally, relying on raw attention maps necessitates brittle heuristics, including layer selection and manual thresholding. We provide a detailed ablation in Appendix~\ref{subsec:supp_decoder_ablation}. 

To resolve these bottlenecks, we employ a lightweight segmentation decoder (\cref{fig:method_training}, top-right). The decoder learns to fuse concatenated multi-layer \textsc{PartConcept} attention maps while being explicitly conditioned on the clean RGB image across multiple spatial scales. This architecture couples the coarse semantic localization encoded in the diffusion attention maps with the sharp, uncorrupted edge cues of the clean image, yielding precise, end-to-end part segmentations.


During part-segmentation inference, given an input image and query part phrases, we first map the phrases into \textsc{PartConcept} tokens (\cref{fig:method_summarizer}) and encode the image into diffusion latents with a fixed noise timestep (details in Appendix~\ref{subsec:supp_inference_timestep}). A single mmDiT forward pass then extracts the \textsc{PartConcept} cross-attention maps, which the decoder processes into final part masks (\cref{fig:method_training}).


\textbf{Preserving the generative capability.} While the localization objective grounds \textsc{PartConcept} tokens to their respective visual regions, directly fine-tuning the model on segmentation alone risks degrading its core synthesis capabilities. To retain the base model's generative prior, we include the flow matching objective (velocity prediction MSE) alongside the localization loss. The joint training objective is formulated as: $\mathcal{L} = \mathcal{L}_\text{dice} + \lambda \cdot \mathcal{L}_\text{flow}$, where $\lambda$ governs the trade-off between localization precision and visual fidelity of generation outputs, as analyzed in~\cref{subsec:ablations}. 


\textbf{Training dataset.} To train our model, we leverage the Pascal-Part~\citep{pascalpart} dataset, chosen for its fine-grained, instance-level spatial annotations of object parts. However, naively training modern diffusion models on Pascal-Part poses a domain alignment challenge: its legacy images feature low resolutions and skewed aspect ratios that can degrade the high-fidelity visual priors of the base T2I model. To overcome this, we design an automated restoration pipeline using FLUX.2 that upsamples and outpaints the images while strictly preserving the underlying part annotations. Training on this enhanced benchmark, termed ``Pascal-Part SR'' (Super-Resolution), ensures compatibility with the visual quality of modern generative models~\citep{flux-2-2025} (see Appendix~\ref{sec:pascal_part_sr} for details).

\textbf{Attribute binding.} Beyond strong attention localization, accurate \textit{part-level instruction following} requires that the \textsc{PartConcept} tokens semantically compose well with surrounding textual attributes. For instance, in a prompt specifying distinct textures or colors for the `right upper arm' versus `left lower arm' (Fig.~\ref{fig:teaser}c), the model must correctly bind each attribute with its corresponding \textsc{PartConcept} token. Rather than introducing explicit binding mechanisms, we foster compositionality at the data level: we augment training prompts with localized part captions extracted by querying a VLM~\citep{bai2025qwen3} on cropped part regions (see Appendix~\ref{sec:vlm_part_captioning} for details). Remarkably, this prompt augmentation implicitly enables robust attribute-part binding during generation, without any explicit architectural mechanisms. 


\section{Experiments} 

We first evaluate our method's effectiveness on part-level localization using the standard \textit{part segmentation}~\cite{ov-parts} (\cref{subsec:experiments_ovps}) and \textit{part instance segmentation}~\cite{pascalpart} benchmarks (\cref{subsec:pascalpart}). We then show that the proposed \textsc{PartConcept}s' strong localization directly enables fine-grained \textit{part-level instruction following} in T2I generation (\cref{subsec:experiments_t2i}). 


\subsection{Open vocabulary part segmentation (OVPS)~\cite{ov-parts}} 
\label{subsec:experiments_ovps}
\textbf{Implementation and training details.} We use FLUX.2 Klein Base 9B~\cite{flux-2-2025} as the base model for all experiments in this paper. For the \textsc{PartConcept} encoder, we train a LoRA (rank 128) over the text-encoder layers (see~\cref{fig:method_summarizer}). Additionally, we train a smaller LoRA (rank 4) for the mmDiT, following standard practice. We train the models until convergence, at batch size 16, using default optimizer settings from HuggingFace Diffusers~\cite{von-platen-etal-2022-diffusers}. We train our models on the respective train splits from the OVPS benchmark, following standard evaluation protocols.


\begin{table}[t]
    \caption{\textbf{Zero-shot evaluation results.} Our method outperforms dedicated segmentation baselines on Open-Vocabulary Part Segmentation (OVPS)~\cite{ov-parts} benchmark, establishing state-of-the-art.}
    \centering
    \begin{tabular}{lccccccccc}
        \toprule
        \multirow{2}{*}{Method} & \multicolumn{3}{c}{ADE20K ~\cite{ov-parts}} & \multicolumn{3}{c}{PPS-116 ~\cite{ov-parts}} & \multicolumn{3}{c}{PartImageNet ~\cite{he2022partimagenet}} \\
        \cmidrule(lr){2-4} \cmidrule(lr){5-7} \cmidrule(lr){8-10}
        & seen & unseen & h-IoU & seen & unseen & h-IoU & seen & unseen & h-IoU \\
        \midrule
        VLPart ~\cite{vlpart}         & * & * & * & 42.6 & 18.7 & 26.0 & * & * & * \\
        ZSSeg+ ~\cite{zsseg}         & 43.2 & 27.8 & 33.9 & 54.4 & 19.0 & 28.2 & * & * & * \\
        CLIPSeg ~\cite{clipseg}       & 38.2 & 30.9 & 34.2 & 48.9 & 27.5 & 35.2 & 53.9 & 37.2 & 44.0 \\
        CAT-Seg ~\cite{catseg}         & 33.8 & 25.9 & 29.3 & 43.8 & 27.7 & 33.9 & 47.3 & 35.1 & 40.3 \\
        PartCLIPSeg ~\cite{partclipseg}    & 38.4 & 38.8 & 38.6 & 50.0 & 31.7 & 38.8 & 56.3 & 51.7 & 53.9 \\
        PartGLEE ~\cite{li2024partglee}        & 51.3 & 35.3 & 41.8 & 57.4 & 27.4 & 37.1 & * & * & * \\
        PartCATSeg ~\cite{partcatseg}     & 53.1 & 47.2 & 50.0 & 57.5 & 44.9 & 50.4 & 73.8 & 71.5 & 72.7 \\
        LangHOPS ~\cite{miao2025langhops}        & 49.3 & 49.7 & 49.5 & 59.2 & 46.5 & 52.1 & 71.9 & \textbf{73.7} & 72.8 \\
        \textbf{Ours} & \textbf{53.4} & \textbf{61.9} & \textbf{57.3} & \textbf{65.0} & \textbf{49.7} & \textbf{56.3} & \textbf{75.4} & 70.6 & \textbf{72.9} \\
        \bottomrule
    \end{tabular}
    \label{tab:zero_shot_ovps}
    \vspace{-5mm} 
\end{table}

\begin{wraptable}{r}{0.6\textwidth} 
    
    \vspace{-2mm}
    \caption{\textbf{Cross-dataset evaluation results.} We consider two settings: PartImageNet (OOD)~\cite{he2022partimagenet} and PartImageNet (train) $\rightarrow$ PPS-116~\cite{ov-parts} (eval).}
    \centering
    \begin{tabular}{lcc}
        \toprule
        \multirow{3.5}{*}{Method} & PartImageNet & PartImageNet $\rightarrow$ \\
        & (OOD)~\cite{he2022partimagenet} & PPS-116 \cite{ov-parts} \\
        \cmidrule(lr){2-2} \cmidrule(lr){3-3}
        & Unseen & Unseen \\
        \midrule
        CLIPSeg ~\cite{clipseg} & 55.86 & 14.87 \\
        CAT-Seg ~\cite{catseg}  & 39.12 & 12.50 \\
        PartCLIPSeg ~\cite{partclipseg} & 59.16 & 19.86 \\
        PartCATSeg ~\cite{partcatseg} & 66.15 & 22.88 \\
        \textbf{Ours} & \textbf{67.30} & \textbf{30.00} \\
        \bottomrule
    \end{tabular}
    \label{tab:cross_dataset}
    \vspace{-2mm}
\end{wraptable}

\textbf{Datasets and evaluation.} Following OVPS~\cite{ov-parts}, we evaluate two setups: \textit{zero-shot}, where train and test categories do not overlap, and \textit{cross-dataset}, where evaluation is performed on a dataset different than the one used for training, hence testing OOD generalization. In the context of \textit{zero-shot} part segmentation, we use Pascal-Part-116 (PPS-116)~\cite{ov-parts}, ADE20K-Part-234~\cite{ov-parts} and PartImageNet~\cite{he2022partimagenet} datasets. Following~\cite{choi2025fine}, we evaluate \textit{cross-dataset} transfer for 1) PartImageNet (training) to PPS-116 (evaluation), and 2) PartImageNet (OOD), a split of the PartImageNet dataset which was proposed originally for few-shot part segmentation~\cite{he2022partimagenet}. Similar to~\cite{miao2025langhops}, we use the oracle-object setting~\cite{ov-parts}, where ground truth object segmentation and class are provided. Following prior work~\cite{ov-parts,choi2024understanding,choi2025fine,miao2025langhops}, we use the IoU metric, reported for both seen and unseen categories separately, as well as their harmonic mean (h-IoU). Further details are presented in Appendix~\ref{sec:ovps_dataset_details}.


\textbf{Baseline comparison.} We compare against standard part segmentation baselines; recent works include PartCATSeg~\cite{choi2025fine} and LangHOPS~\cite{miao2025langhops}. Evaluation results  are presented in~\cref{tab:zero_shot_ovps} (\textit{zero-shot}) and ~\cref{tab:cross_dataset} (\textit{cross-dataset}). Our method outperforms dedicated segmentation baselines in almost all categories, establishing state-of-the-art. We visualize segmentation results in~\cref{fig:experiments_ovps_qual_compare}.

\begin{wraptable}{r}{0.50\textwidth}
    \caption{\textbf{Fine-grained part \textit{instance} segmentation}, with spatial identification of part instances (e.g. `\underline{left lower} arm', `\underline{right upper} arm').}
    \centering
    \begin{tabular}{lccc}
        \toprule
        \multirow{2}{*}{Method} & \multicolumn{3}{c}{Pascal-Part (org.)~\cite{pascalpart}} \\
        \cmidrule(lr){2-4}
        & seen & unseen & h-IoU \\
        \midrule
        PartCATSeg ~\cite{partcatseg} & 33.2 & 35.5 & 34.3 \\
        LangHOPS ~\cite{miao2025langhops}   & 38.4 & 38.7 & 38.5 \\
        \textbf{Ours} & \textbf{55.2} & \textbf{54.2} & \textbf{54.7} \\
        \bottomrule
    \end{tabular}
    \label{tab:pp_spatial}
\end{wraptable}

\subsection{Fine-grained part \textit{instance} segmentation~\cite{pascalpart}} 
\label{subsec:pascalpart}

\begin{figure}[t] 
  \centering
  \includegraphics[width=0.9\linewidth]{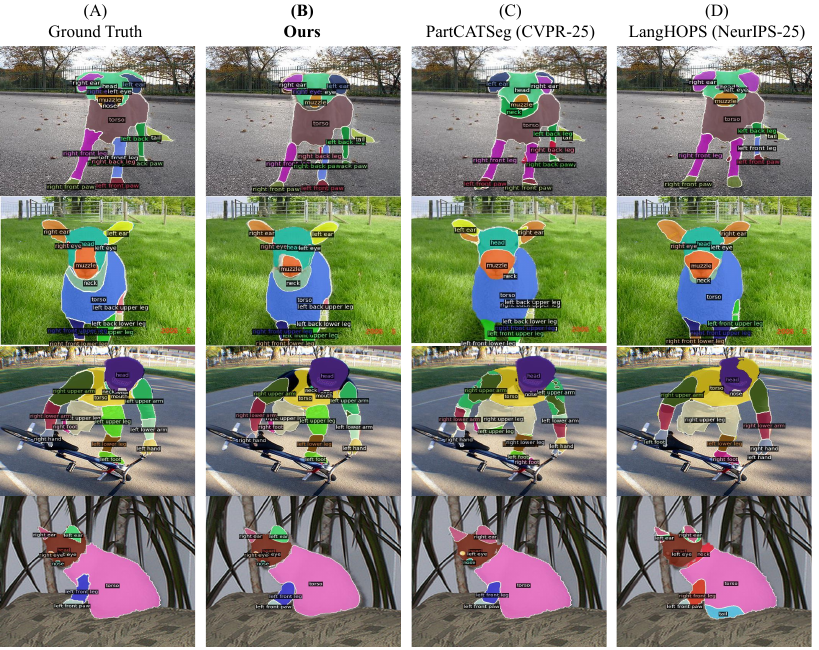} 
  \vspace{-1mm}
  \caption{\textbf{Qualitative comparison.} Our method outperforms all baselines on various benchmarks, establishing state-of-the-art. Notably, we are significantly better for \textit{part instance segmentation}~\cite{pascalpart}, effectively disambiguating spatially distinct part instances like `\underline{left lower} arm' and `\underline{right upper} arm'.} 
  \vspace{-5mm}
\label{fig:experiments_ovps_qual_compare} 
\end{figure}

While our method is competitive on (OVPS)~\citep{ov-parts} benchmark, we find that the OVPS suite is too coarse for precise grounding: it typically removes spatial descriptors and collapses distinct instances such as `left hand' and `right hand' into a single `hand' instance (see~\cref{fig:ovps_vs_pascalpart} and Appendix~\ref{sec:ovps_vs_pascalpart}). Therefore, we evaluate for part \textit{instance} segmentation by training our method (and baselines) on the original Pascal-Part~\citep{pascalpart} benchmark, which retains these fine-grained part instances and corresponding spatial descriptions. On this more rigorous benchmark, our method significantly outperforms dedicated perception baselines and establishes a new state-of-the-art. We visualize segmentation results in~\cref{fig:experiments_ovps_qual_compare}. 

\begin{figure}[t] 
  \centering
  \includegraphics[width=\textwidth]{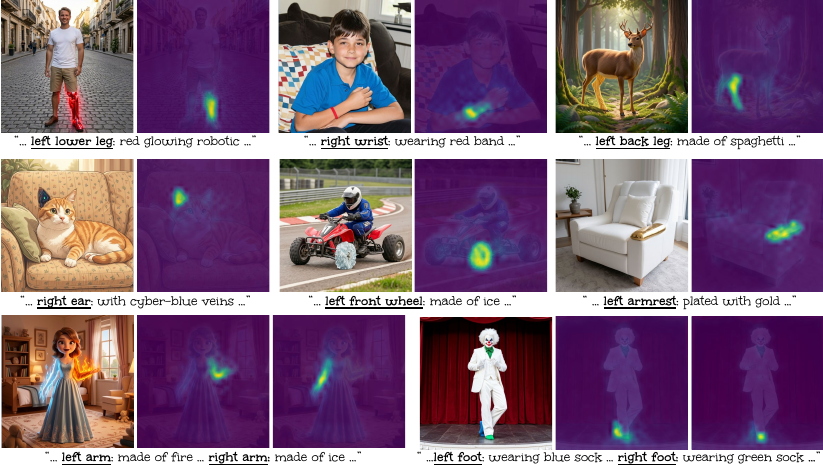} 
  \caption{\textbf{Qualitative results for \textit{part-level instruction following}, with \textsc{PartConcept} attention maps.} Our method is able to disambiguate spatial identifiers in part descriptions, (e.g., `\underline{left front} wheel'). Notably, \textsc{PartConcept}s generalize to novel part phrases; specifically, \textbf{\textit{armrest, left front wheel} and \textit{wrist} are not seen in our training data}, indicating strong generalization. Even in case of multiple parts, the attributes are correctly bound to respective \textsc{PartConcept}s; this demonstrates that \textsc{PartConcept} embeddings preserve the compositional nature of text embeddings.} 
  \label{fig:experiments_qual_ours_with_localization} 
  \vspace{-6mm}
\end{figure}

\begin{wrapfigure}{r}{0.57\textwidth} 
  \vspace{-10mm} 
  \centering
  \includegraphics[width=\linewidth]{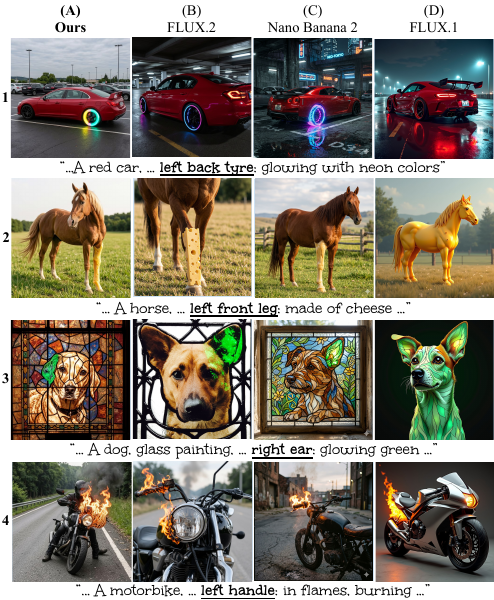} 
  \caption{\textbf{Qualitative comparisons.} Our method exhibits strong spatial localization, while baselines frequently struggle, failing to disambiguate spatial identifiers in part descriptions (e.g., `\underline{left back} wheel', `\underline{right} ear', etc.).}  
  \label{fig:experiments_qual_compare}
  \vspace{-7mm}
\end{wrapfigure}
\vspace{-1mm}
\subsection{Fine-grained instruction following} 
\label{subsec:experiments_t2i} 

\textbf{Implementation and training details.} We use the same settings as described in~\cref{subsec:experiments_ovps}. Additionally, to preserve the generative capability of the base T2I model, we set the loss balancing coefficient $\lambda = 0.25$ (~\cref{sec:method}). We train this model on the Pascal-Part SR dataset, thus preserving the photorealistic prior of the generative model. 

\textbf{Baselines.} We compare our method against our base T2I model FLUX.2 Klein Base 9B~\cite{flux-2-2025}, Nano Banana 2~\cite{nanobanana2,team2023gemini}, and FLUX.1 [dev]~\cite{labs2025flux}. 

\textbf{Quantitative evaluation.} We primarily evaluate our method (and baselines) for \textbf{attribute localization}, i.e., \textit{how strongly are the attributes bound to the intended parts}? We construct an evaluation dataset of part-level instructions: given a data-point from the Pascal-Part dataset, we sample a part label and couple it with a randomized instruction, such as `stain with purple color'. The resulting text prompts follow the template `a photo of \texttt{\{class\_name\}}, with \texttt{\{instruction\}}'. We construct evaluation examples with one and two part-level instructions per prompt. Part-masks for generated images are obtained using our \textsc{PartConcepts} segmentation model (\cref{subsec:experiments_ovps}), followed by manual refinement. We use these part-masks to evaluate localized attribute correctness using a VLM-as-a-judge framework, which has been used for evaluating compositional T2I generation in prior work~\cite{li2024genai,huang2025t2i}. Specifically, the VLM is provided with an annotated image with the target part boundary highlighted, and asked to provide a score according to the following rubric: 

\vspace{-1mm}
\begin{itemize} 
\item 1.0 if the attribute is not present in indicated region.
\item 2.0 to 4.0 if the attribute is present in the indicated region, but also leaks to other parts/regions outside the part boundary. A score of 2.0 indicates severe leakage onto other parts, while 4.0 indicates minor/slight leakage.
\item 5.0 if the attribute is present in the indicated part AND is localized well.
\item A score of 0.0 is automatically assigned if the part is not visible in the generated image (empty-mask). 
\end{itemize}
\vspace{-1mm}

\begin{wrapfigure}{r}{0.30\textwidth}
  \vspace{-5mm}
  \centering
  \includegraphics[width=\linewidth]{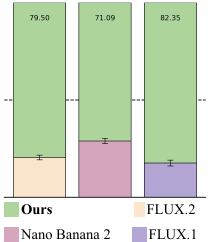}
  \caption{\textbf{User study} evaluating localization of attributes to respective parts.}
  \label{fig:user_study}
  \vspace{-6mm}
\end{wrapfigure}

We perform this evaluation using Qwen3-VL-8B-Instruct~\cite{bai2025qwen3} and Gemini 3.5 Flash~\cite{team2023gemini}, with results reported in~\cref{tab:generation_vlm_scores}. FLUX.1 [dev] applied the requested attribute to the entire object for most cases (see~\cref{fig:experiments_qual_compare} D2-3), and hence we removed it from this evaluation. We also performed an A/B user study (see~\cref{fig:user_study}), where users were asked to compare between our method's output and a randomly chosen baseline using similar instructions as the above rubric. More details about the user study are provided in Appendix \ref{section:supp_user_study}.
The evaluations consistently highlight that our method enables more accurate part-level instruction following compared to baselines.


\textbf{Qualitative evaluation.} As shown in~\cref{fig:experiments_qual_ours_with_localization}, the \textsc{PartConcept} strongly localizes to correct part regions, shown by the sharp attention maps, while preserving T2I generation fidelity. Our \textsc{PartConcept} encoder exhibits strong generalization to novel part phrases (e.g., \textbf{\textit{armrest}, \textit{left front wheel}, \textit{wrist} (\cref{fig:experiments_qual_ours_with_localization}) and \textit{handle}, \textit{tyre} (\cref{fig:experiments_qual_compare}) are not present in our training data}).

\begin{wraptable}{l}{0.575\textwidth}
  \centering
  \caption{\textbf{VLM-as-judge} evaluation of attribute localization.}
  \label{tab:generation_vlm_scores}
  \begin{tabular}{lcc}
      \toprule
      Model & Qwen~\cite{bai2025qwen3} & Gemini~\cite{team2023gemini} \\
      \midrule
      Nano Banana 2~\cite{nanobanana2,team2023gemini} & 2.49 & 1.73 \\
      FLUX.2 Klein~\cite{flux-2-2025} & 2.34 & 1.60 \\
      \textbf{Ours} & \textbf{3.52} & \textbf{2.45} \\
      \bottomrule
  \end{tabular}
\end{wraptable}

Further, our method effectively leverages the T2I generative prior to generalize to arbitrary attributes: none of the shown part attributes in~\cref{fig:experiments_qual_ours_with_localization,fig:experiments_qual_compare} were seen in our training data. Notably, our method performs correct attribute binding even in multi-part scenarios (third row,~\cref{fig:experiments_qual_ours_with_localization}), without any explicit mechanisms for attribute binding. This demonstrates that the \textsc{PartConcept} token preserves compositional nature of the text space, and thus behaves \textit{similarly} to other text tokens. Fig.~\ref{fig:experiments_qual_compare} (B,C) shows that FLUX.2 and Nano Banana 2 often fail to generate the intended part in some cases, or fail to correctly bind the attribute. In FLUX.1 generations the attribute is highly diffused, and not restricted to small part regions (\cref{fig:experiments_qual_compare} D).


\subsection{Ablations} 
\label{subsec:ablations}

\begin{wrapfigure}{r}{0.37866\textwidth}
    
  \centering
  \vspace{-17mm}

  \includegraphics[width=\linewidth]{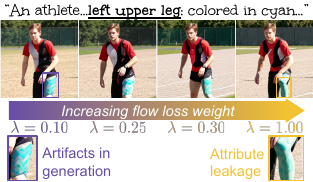}
  \captionof{figure}{\textbf{Effect of increasing flow loss weight ($\lambda)$}. Low values lead to visual artifacts (left), while higher values cause attribute leakage.}
  \label{fig:flow_loss_weight}

  \includegraphics[width=\linewidth]{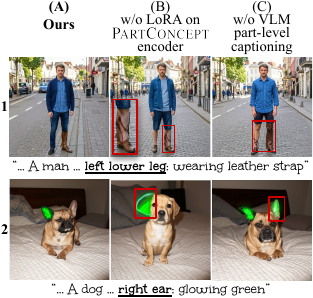}
  \captionof{figure}{\textbf{Ablations.} \textbf{B.} LoRA on the \textsc{PartConcept} is crucial for preserving generation fidelity. \textbf{C.} Training prompt augmentation using VLM captioning improves attribute binding.}
  \label{fig:ablations_main}
  
  \includegraphics[width=\linewidth]{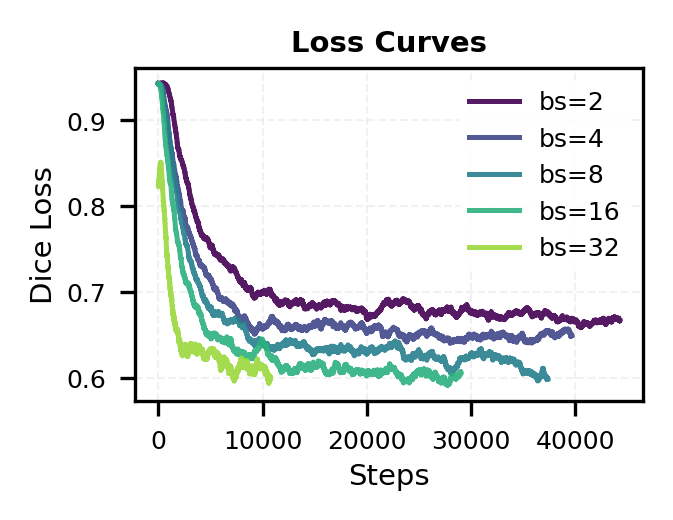}
  \vspace{-8mm}
  \captionof{figure}{\textbf{Scaling batch size.} The localization loss trajectories respond positively to increasing batch sizes.}
  \vspace{-10mm} 
  \label{fig:batch_size_scaling}
\vspace{-10mm}
\end{wrapfigure}


We validate our approach through several ablative experiments. First, the flow loss weight ($\lambda$) balances localization and denoising objectives during training (see~\cref{fig:flow_loss_weight}). We find that $\lambda$ dictates a tradeoff between generation fidelity and attribute binding (see~\cref{fig:flow_loss_weight}); low values degrade the generative prior, causing visual artifacts, while high values reduce localization pressure, leading to attribute leakage. \\  
Second, LoRA finetuning for \textsc{PartConcept} encoder (see~\cref{fig:method_summarizer}) is critical; full fine-tuning causes significant visual artifacts (highlighted with \textcolor{red}{red} boxes in~\cref{fig:ablations_main}(B)), as \textsc{PartConcept} embedding drifts too much from the text space, disrupting the T2I prior. Third, our VLM-based prompt augmentation is essential for attribute binding; without it, the training lacks incentive to associate specific attributes with \textsc{PartConcept}s, as shown in~\cref{fig:ablations_main}(C). \\ 
Finally, we find batch size significantly impacts localization performance. Unlike standard FLUX.2~\cite{flux-2-2025} fine-tuning scripts~\cite{von-platen-etal-2022-diffusers} that use small batches, our localization loss trajectories significantly improve with larger batch sizes (\cref{fig:batch_size_scaling}, we also expect improvements with scaling). We further ablate upon additional components of the trained \textsc{PartConcepts} model in Appendix~\ref{section:supp_ablation_study}. 

\section{Limitations}
\label{sec:limitations}
Despite its effectiveness, \textsc{PartConcepts} faces two primary limitations. First, scaling beyond four part-attribute pairs occasionally causes attribute leakage (see Appendix~\ref{sec:failure_example}), even when spatial localization is accurate. This likely stems from a distributional shift, as prompts with more than four attributes are out-of-distribution (OOD) relative to our training data. Second, while we provide a unified mechanism for segmentation and T2I part-level control, the best-performing loss weights ($\lambda$) differ for each task. This reflects a slight trade-off between discriminative grounding and generative prior preservation. Future research should try to mitigate this gap by designing the model in a way that it leads to optimal discriminative localization capabilities without having any tradeoff on generative prior preservation.
\section{Conclusion} 
\label{sec:conclusion}
This work stems from observing that while T2I cross attention maps are strongly disentangled for objects, this semantic clarity collapses at the part-level, particularly for spatially distinct part instances (e.g., `left front leg' vs. `right front leg').  We identify the bottleneck as text-image interaction (cross attention) rather than the visual feature resolution. Hence, we propose a lightweight mechanism to correct this: encoding part phrases (e.g. `left front leg') into a compact representation: the \textsc{PartConcept} tokens, trained for spatial grounding. Our approach effectively leverages the base T2I's generative priors to achieve state-of-the-art performance on various \textit{part (instance) segmentation} benchmarks, outperforming dedicated perception baselines. We further show that this effective localization directly enables robust attribute binding and \textit{fine-grained instruction following} in T2I generation. These results demonstrate the effectiveness of the proposed approach for improving part-level understanding of generative models. We believe that the proposed attention based supervision can be extended as a general post-training mechanism for T2I diffusion models to strengthen specific textual concepts underrepresented during pretraining. We leave this exploration for future work. 

\section{Acknowledgements}
We thank Medha Sen, Amal Joseph and other members of CVIT, IIIT Hyderabad for generously prioritizing compute access for this project, especially during the last days before deadline. We are grateful to Aishwarya Agarwal, Ayan Kashyap, and Rishubh Parihar for helpful discussions, and to Darshana and Medha for help in reviewing the manuscript. This project was supported by a grant from Adobe Research.

\bibliographystyle{unsrtnat} 
\bibliography{main}

\clearpage
\appendix

\section{Analyzing the trained \textsc{PartConcepts} model}
\label{section:supp_ablation_study}

In this section we analyze our trained \textsc{PartConcepts} model, studying the impact of each trainable component (the segmentation decoder, \textsc{PartConcept} encoder and LoRA~\cite{hu2022lora} on the mmDiT), choice of inference-time noise levels for localization inference, and conclude with a discussion about how our approach significantly enhances the capabilities of the base T2I model.

\subsection{Role of the segmentation decoder}
\label{subsec:supp_decoder_ablation}
Directly using \textsc{PartConcept} attention maps for \textit{part instance segmentation} is challenging for two reasons. First, the attention maps in mmDiT are computed using the compressed latents, hence they lack the resolution required for dense part mask prediction at the original image resolution. Second, directly using attention maps for downstream segmentation is quite brittle, as it necessitates extra hyperparameters for attention map thresholding and layer selection.  

\begin{figure}[ht!]
  \centering
  \includegraphics[width=\linewidth]{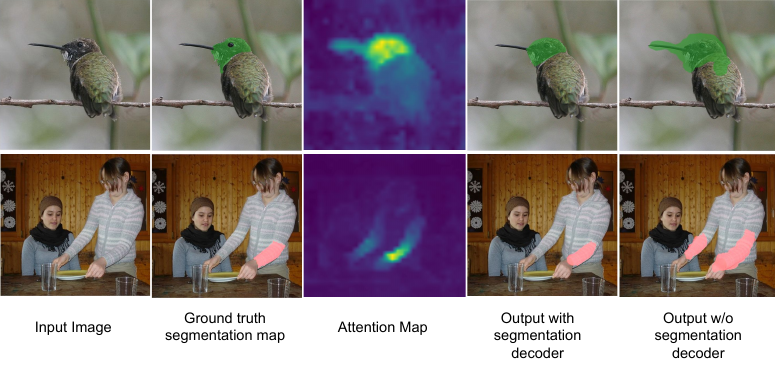} 
  \caption{\textbf{Impact of segmentation decoder on \textit{part instance segmentation}.} While \textsc{PartConcept} attention maps are able to coarsely localize the part regions, using them directly for segmentation fails to match the performance of the full model with segmentation decoder. This is because while the \textsc{PartConcept} attention maps are coarsely localized to the correct region, they lack the fine-grained resolution necessary for dense segmentation prediction; segmentation decoder mitigates this issue by combining the coarse localization of \textsc{PartConcept} attention maps with fine details from the input image to output precise part segmentation maps.} 
  \label{fig:supplementary_segdecoder_nosegdecoder}  
\end{figure}

To demonstrate this, we set up a simple baseline, where we average the layer-wise attention maps obtained after forward pass, average them and apply bilinear upsampling to get a single map, of the same dimensions as the input image. Finally, we apply the adaptive OTSU thresholding~\cite{otsu1975threshold} to get the part segmentation masks. As illustrated in~\cref{fig:supplementary_segdecoder_nosegdecoder}, while this baseline is able coarsely localize the part regions, it fails to match the performance of the full model with the segmentation decoder. This is because, while the \textsc{PartConcept} attention maps are coarsely localized to the correct region, they lack the fine-grained resolution necessary for dense segmentation prediction; the segmentation decoder mitigates this issue by combining the coarse localization of \textsc{PartConcept} attention maps with the fine details from input image to output precise part segmentation maps. 

\subsection{Impact of inference timestep} 
\label{subsec:supp_inference_timestep}
For inferring part segmentation maps, we encode the input image into diffusion latent space and add a fixed amount of noise to it. We investigate the sensitivity of our framework to the noise timestep $t$ used during inference ($t = 0$ is zero noise (clean image), $t = 1000$ is maximum noise). We perform this analysis on the OVPS Pascal-Part split~\cite{ov-parts}. As shown in ~\cref{tab:timestep_ablation}, the model exhibits a stable performance across the mid-range of the noise schedule ($t=200$ to $t=600$), with best performance at $t = 400$. 

\begin{figure}[t] 
  \centering
  \includegraphics[width=\linewidth]{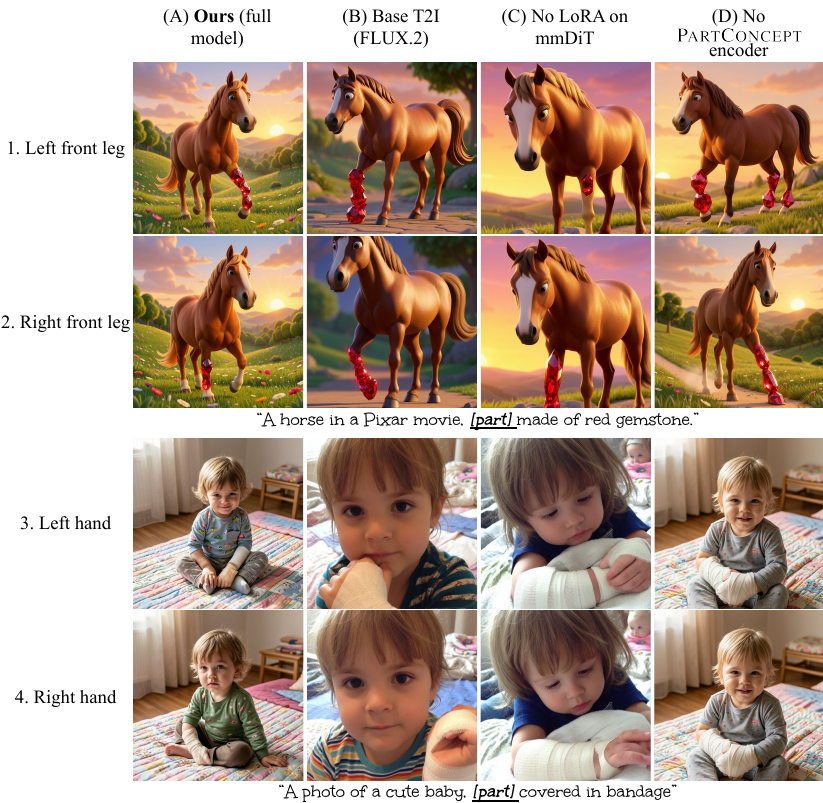} 
  \caption{\textbf{Comparison with base T2I model, and impact of each trainable component on the generative task.} We present A. our method's output, B. base T2I model (FLUX.2) outputs, C. output without LoRA on mmDiT (\cref{subsec:supp_ablation_lora_mmdit}), and D. output without having the \textsc{PartConcept} encoder (\cref{subsec:supp_partconcept_ablation}). We find that the base T2I model (FLUX.2 Klein 9B~\cite{flux-2-2025}) demonstrates lack of spatial part instance understanding, demonstrated by inability to correctly follow part-level instructions (B); our approach significantly enhances this understanding, thus enables finer text-based generative control (\cref{subsec:enhancing_base_t2i}). We find that while removing the LoRA on mmDiT does not have a significant impact (C), \textsc{PartConcept} encoder is crucial for binding the attributes to correct part regions (D). \textit{\textbf{All comparisons presented in this figure use the same starting latent.}}} 
\label{fig:supp_generative_ablation} 
\end{figure}

\begin{table}[ht]
  \caption{\textbf{Impact of the inference timestep $t$ on segmentation performance.} The model exhibits a stable performance across the mid-range of the noise schedule ($t=200$ to $t=600$), with best performance at $t = 400$.}
  \label{tab:timestep_ablation}
  \centering
  \begin{tabular}{lccc}
    \toprule
    \textbf{Timestep ($t$)} & \textbf{mIoU Base} & \textbf{mIoU Novel} & \textbf{hIoU} \\
    \midrule
    200 & 63.90 & 49.19 & 55.50 \\
    400 & \textbf{65.04} & \textbf{49.70} & \textbf{56.30} \\
    600 & 64.87 & 49.47 & 56.10 \\
    800 & 61.88 & 45.42 & 52.30 \\
    \bottomrule
  \end{tabular}
\end{table}

\subsection{Impact of \textsc{PartConcept} encoder}
\label{subsec:supp_partconcept_ablation}
As shown in~\cref{fig:supp_generative_ablation}(D), the \textsc{PartConcept} encoder is essential for \textit{part-level instruction following} in T2I generation; without it, the model fails to bind attributes to the correct part region. This demonstrates that \textsc{PartConcept} encoder is the major bridge that \textit{aligns} the textual part-level phrases to the visual features of the diffusion model. 

\subsection{Role of the LoRA on mmDiT}
\label{subsec:supp_ablation_lora_mmdit}
\begin{figure}[ht!]
  \centering
  \includegraphics[width=\linewidth]{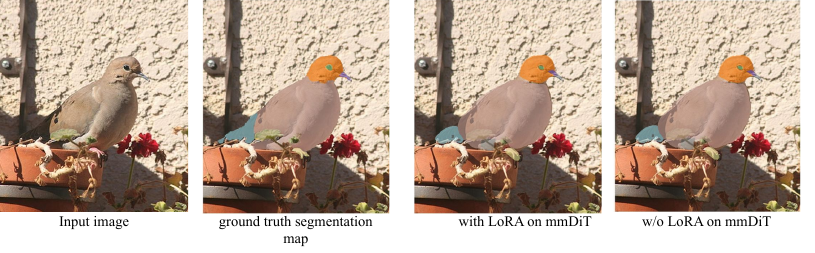} 
  \caption{\textbf{Impact of trainable mmDiT LoRA parameters on \textit{part instance segmentation}.} We find that the mmDiT LoRA parameters have minimal impact on the final part segmentation accuracy.}  
  \label{fig:supplementary_lora_nolora}
\end{figure}

Following standard finetuning scripts provided in the HuggingFace Diffusers~\cite{von-platen-etal-2022-diffusers} framework, we use a small LoRA~\cite{hu2022lora} of rank 4 on the mmDiT. However, as illustrated in~\cref{fig:supplementary_lora_nolora}, this trainable component seems to have minimal impact on the final part segmentation maps. Similarly, for the generative task, the model binds attributes correctly to the respective parts, even in absence of LoRA on the mmDiT~\cref{fig:supp_generative_ablation}(C). 

\subsection{Conclusion: enhancing the base T2I model capabilities.}
\label{subsec:enhancing_base_t2i}
To conclude this analysis, we present a comparison with the base T2I model used for \textsc{PartConcepts}, FLUX.2 Klein 9B~\cite{flux-2-2025}, see~\cref{fig:supp_generative_ablation}(B). Clearly, the base T2I model fails to bind attributes to the correct parts: when prompted to apply gemstone texture (\cref{fig:supp_generative_ablation} 1-2B) separately to the `\underline{left front} leg' and `\underline{right front} leg' of the horse, it applies the texture to the right front leg in both cases. Similar case with `\underline{left} hand' and `\underline{right} hand' in~\cref{fig:supp_generative_ablation} 3-4B, where the model fails to associate the bandage attribute to the correct hand. These results indicate an absence of spatial part instance understanding in the base T2I model. Our approach significantly enhances this understanding, hence enables finer text-based generative control.

\section{Pascal-Part SR: Enhancing Pascal-Part~\cite{pascalpart}} 
\label{sec:pascal_part_sr} 
Pascal-Part~\cite{pascalpart} is a state-of-the-art dataset for part instance segmentation. It is especially useful because of its fine-grained labels, with spatial identification of part instances (e.g., `\underline{left front upper} leg', `\underline{right lower} arm', etc.). Due to its granularity, we utilize it in our model training for \textit{fine-grained instruction following} (see~\cref{subsec:experiments_t2i}). However, images of this dataset are low-resolution and have irregular aspect ratios; training on such low quality images affects the high-fidelity generation capabilities of the base T2I model. 

\begin{figure}[t] 
  \centering
  \includegraphics[width=0.91\linewidth]{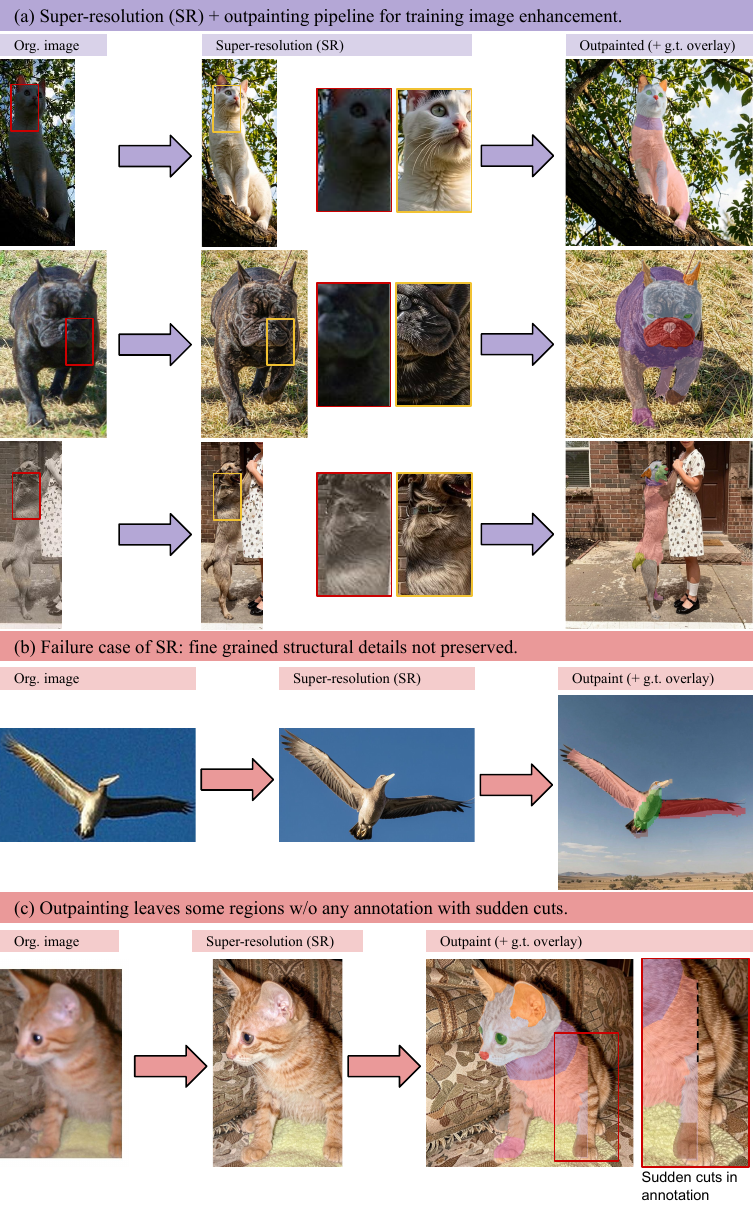} 
  \caption{\textbf{Super-resolution (SR) + outpainting pipeline for training image enhancement.}} 
\label{fig:pascal_part_sr} 
\end{figure}

To mitigate these issues while maintaining the integrity of the original dense labels, we propose an automated image-enhancement pipeline leveraging the restoration capabilities of FLUX.2~\cite{flux-2-2025}. As a pre-processing step, we first isolate individual object instances by cropping them from the original multi-object Pascal-Part images, applying a marginal spatial padding to preserve local context. As illustrated in~\cref{fig:pascal_part_sr}(a), the subsequent pipeline consists of two stages: (1) Super-Resolution (SR) to increase the pixel density of the raw training data, followed by (2) symmetric outpainting to rectify skewed aspect ratios. 

A critical design requirement of this pipeline is the preservation of dense, part-level annotations across both transformations. During the SR stage, which maintains the original aspect ratio, we apply an identical scaling factor to the segmentation masks to ensure pixel-perfect alignment with the upscaled features. In the subsequent outpainting stage, we employ a symmetric padding strategy, extending the canvas boundaries equally relative to the image center. This approach ensures that the original image content—and its corresponding dense annotations—remains undisturbed, allowing for a seamless mapping to the new coordinate space.  

While the proposed pipeline successfully enhances the majority of Pascal-Part~\cite{pascalpart} images, certain edge cases, specifically images with extreme low resolution or severe aspect ratio skew present challenges. \textbf{Structural detail degradation in SR (\cref{fig:pascal_part_sr}(b)):} For images below a certain critical resolution, the super-resolution fails to recover fine-grained structural details, leading to hallucinations. To maintain the high fidelity required for model training, we implement a minimum resolution threshold, filtering out samples that lack sufficient initial information for high-quality restoration. \textbf{Annotation discontinuity in outpainting (\cref{fig:pascal_part_sr}(c)):} Because the outpainting process generates new visual content beyond the original image boundaries, these newly synthesized regions lack corresponding ground-truth labels. This results in "semantic cuts" or abrupt boundaries where the original dense part annotations terminate. We accept this trade-off to achieve the uniform aspect ratios necessary for preserving generative capabilities of the base T2I.  

The resulting Pascal-Part SR dataset consists of 15493 enhanced object-instance images, out of 20173 total valid instances in the original Pascal-Part dataset. Training on this enhanced dataset ensures compatibility with the high visual quality of state-of-the-art generative models~\cite{flux-2-2025}. 

\section{VLM-based part-level captioning}
\label{sec:vlm_part_captioning} 

\begin{figure}[h!] 
  \centering
  \includegraphics[width=0.9\linewidth]{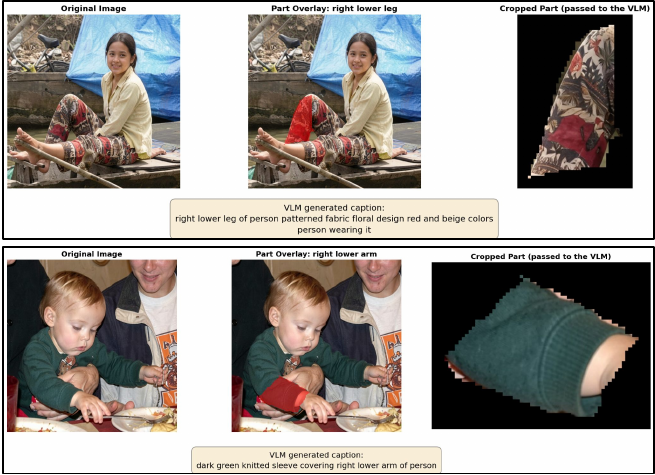}
  \caption{\textbf{VLM-based part captioning.} To ensure that the \textsc{PartConcept} tokens compose well with other text tokens, we augment the training prompts with part-level captions: we pass cropped out part regions from the training images through a VLM~\cite{yang2025qwen3}, which generates suitable captions describing the part regions.}  
  \label{fig:vlm_part_captioning} 
\end{figure}

To ensure that the \textsc{PartConcept} tokens compose well with other text tokens, we augment the training prompts with part-level captions: we pass cropped out part regions from the training images through a VLM~\cite{yang2025qwen3}, which generates suitable captions describing those parts (see~\cref{fig:vlm_part_captioning}).

\section{Failure case: dense text prompts w/ many parts}  
\label{sec:failure_example}
\begin{figure}[t] 
  \centering
  \includegraphics[width=\linewidth]{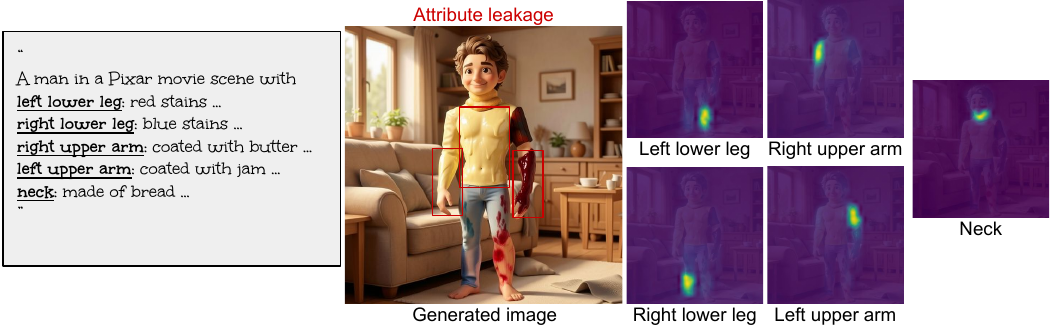} 
  \caption{\textbf{Example failure case.} Dense text prompts with many parts cause attribute leakage.}  
\label{fig:failure_example} 
\end{figure}

Despite its effectiveness, \textsc{PartConcepts} has a limitation: scaling beyond four part-attribute pairs occasionally causes attribute leakage, even when spatial localization is accurate. This stems from a distributional shift, as high-density prompts are out-of-distribution (OOD) relative to our training data. An example is shown in~\cref{fig:failure_example}.  

\section{Compute resources}
\label{sec:compute_resources}
Our training setup consists of 2 NVIDIA H100 GPUs (96G VRAM each) with one INTEL(R) XEON(R) GOLD 6526Y CPU and 512G memory. Most of our training runs and ablations take roughly 24 hours on this setup. 

\section{Additional qualitative comparisons} 
\begin{figure}[htb!] 
  \centering
  \includegraphics{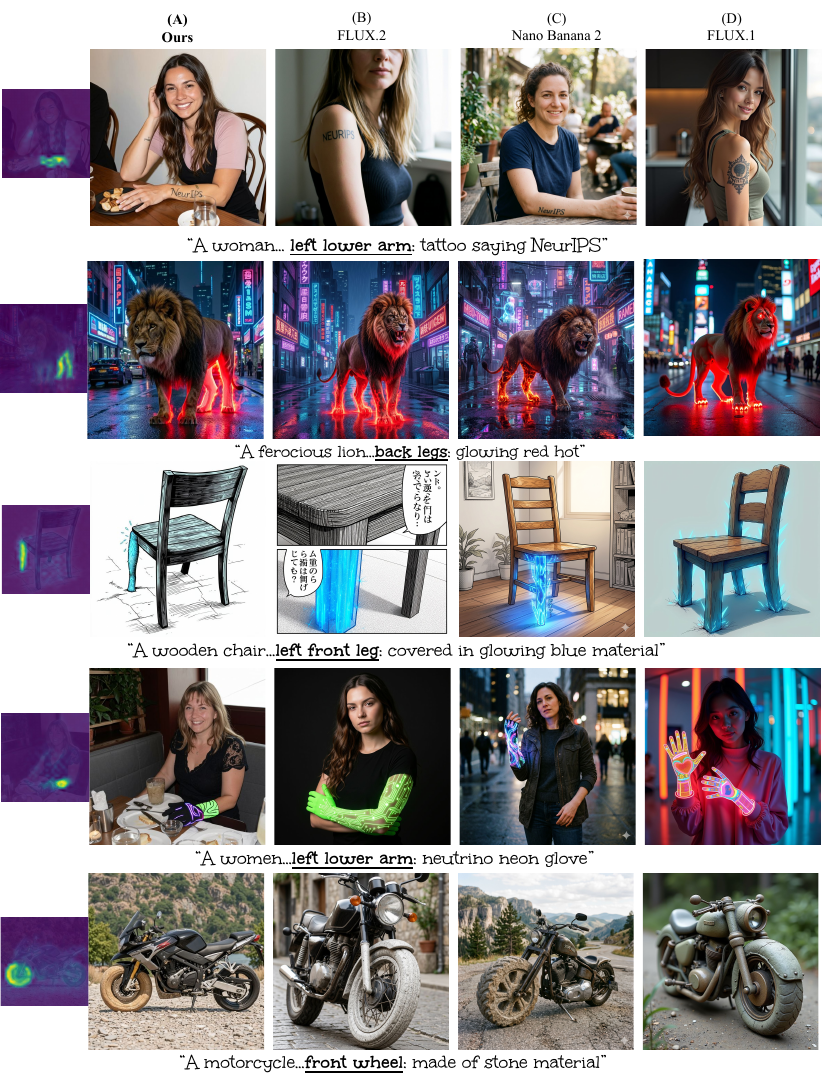}
  \caption{\textbf{Qualitative comparisons.} Our method exhibits strong spatial localization, while baselines frequently struggle, failing to disambiguate spatial identifiers in part descriptions (e.g., `\underline{left front} leg', `\underline{back} legs', `\underline{left lower arm}' etc.).} 
  \label{fig:supp_qual_baselines_1} 
\end{figure}

\begin{figure}[htb!] 
  \centering
  \includegraphics{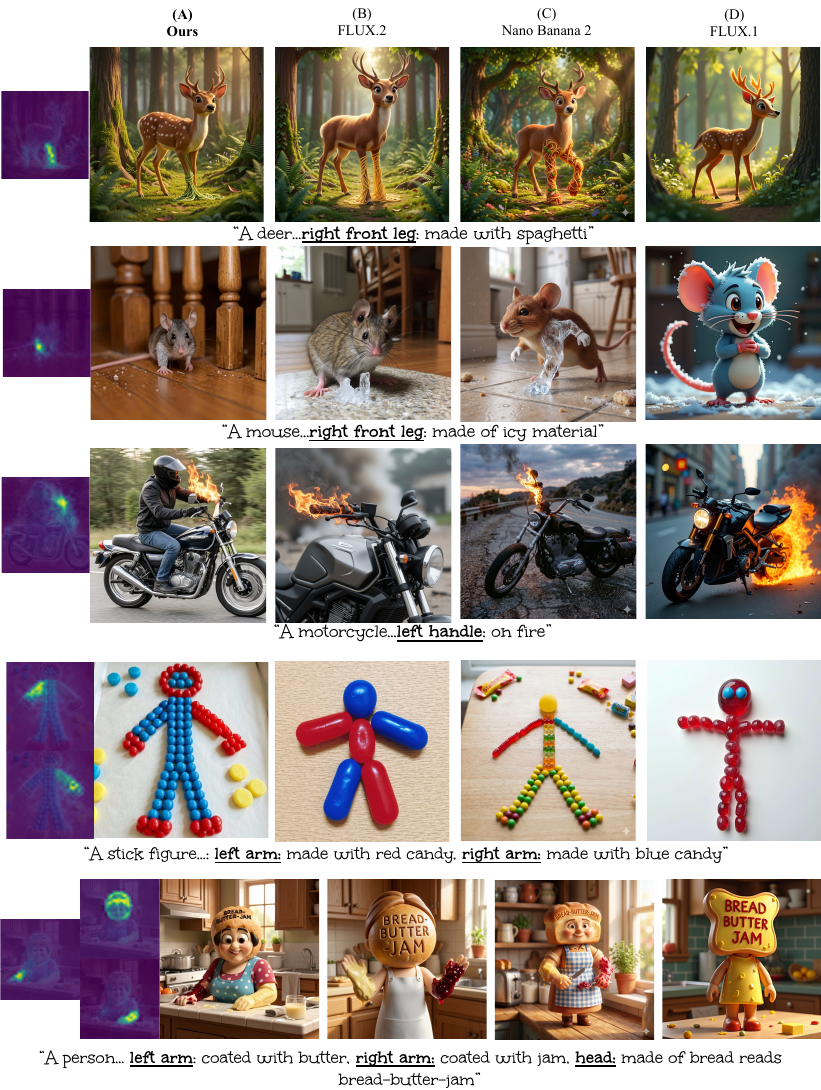}
  \caption{\textbf{Qualitative comparisons.} Our method exhibits strong spatial localization, while baselines frequently struggle, failing to disambiguate spatial identifiers in part descriptions (e.g., `\underline{right front} leg', `\underline{left} handle', `\underline{right} arm' etc.).}
  \label{fig:supp_qualitative_baselines_2} 
\end{figure}

\begin{figure}[ht!] 
  \centering
  \includegraphics{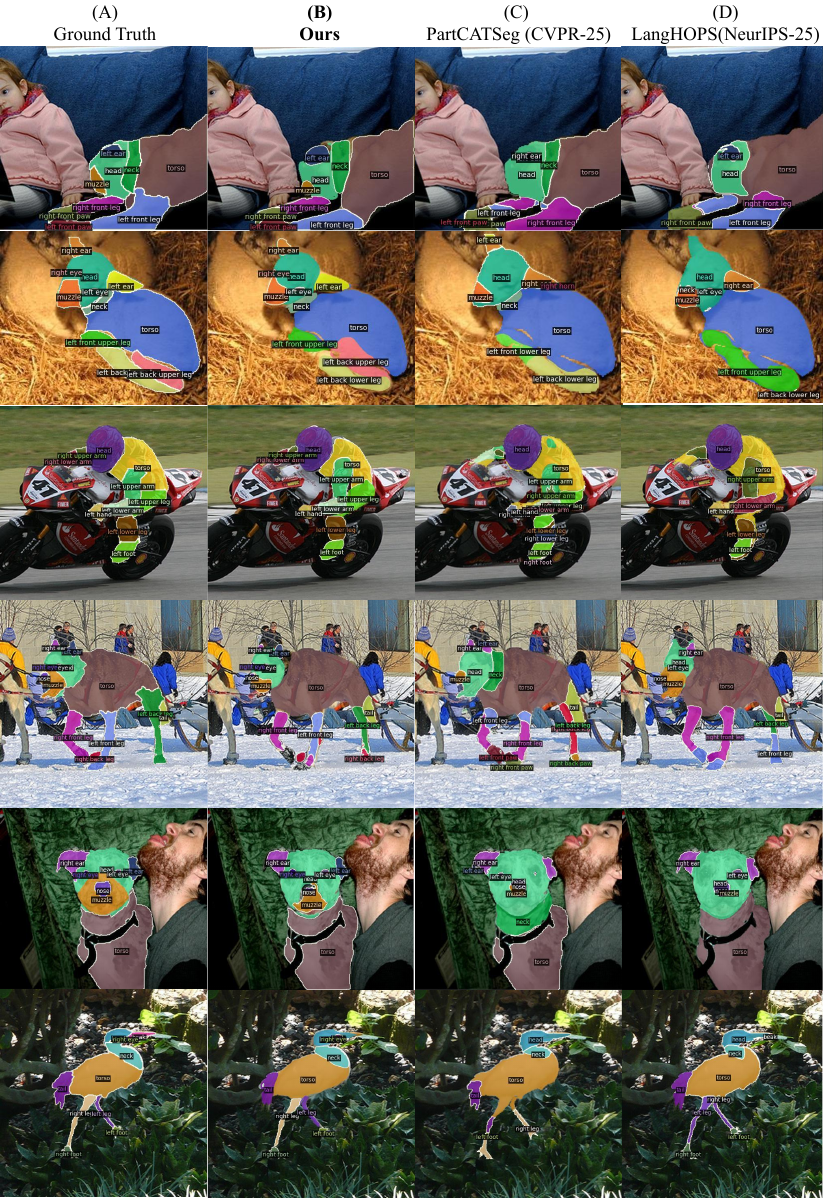} 
  \caption{\textbf{Qualitative comparison.} Our method outperforms all baselines on various benchmarks, establishing state-of-the-art. Notably, we are significantly better for \textit{part instance segmentation}~\cite{pascalpart}, effectively disambiguating spatially distinct part instances like `\underline{left lower} arm' and `\underline{right upper} arm'.} 
  \label{fig:supp_qual_compare_segmentation} 
\end{figure}

We present qualitative comparisons for \textit{fine-grained instruction following} in T2I generation (\cref{fig:supp_qual_baselines_1,fig:supp_qualitative_baselines_2}) and \textit{part instance segmentation} (\cref{fig:supp_qual_compare_segmentation}).

\clearpage 

\clearpage 

\clearpage

\section{Details about Open-Vocabulary Part Segmentation (OVPS)~\cite{ov-parts} benchmark}
\label{sec:ovps_dataset_details}

\textbf{Pascal-Part-116.} The Pascal-Part-116 dataset serves as an adaptation of the original Pascal-Part dataset~\cite{pascalpart}, tailored specifically to support Open-Vocabulary Part Segmentation (OVPS) research. It incorporates a combination of base and novel object classes to ensure variety. Notably, the object categories ``bird'',  ``car'', ``dog'', ``sheep'' and ``motorbike'' are explicitly treated as novel. At the part level, the dataset is structured with 74 base and 42 novel object-specific part categories, establishing a robust framework for evaluating segmentation models. Additional details regarding the dataset are given in~\cref{tab:pascal_part_116_enclosed}. 

\begin{xltabular}{\textwidth}{@{} l X @{}}
    \caption{Pascal-Part-116 base, novel categories and their associated part labels.} \label{tab:pascal_part_116_enclosed} \\
    \toprule
    \textbf{Object} & \textbf{Parts} \\
    \midrule
    \endfirsthead

    \toprule
    \textbf{Object} & \textbf{Parts} \\
    \midrule
    \endhead

    \midrule
    \multicolumn{2}{r}{{Continued on next page\ldots}} \\
    \bottomrule
    \endfoot

    \bottomrule
    \endlastfoot

    \multicolumn{2}{c}{\textit{Novel Categories (42 Parts)}} \\
    \midrule
    Bird        & beak, eye, foot, head, leg, neck, tail, torso, wing \\
    \addlinespace
    Car         & back, door, front, headlight, license plate, mirror, roof, side, wheel, window \\
    \addlinespace
    Dog         & ear, eye, head, leg, muzzle, neck, nose, paw, tail, torso \\
    \addlinespace
    Motorbike   & handlebar, headlight, saddle, wheel \\
    \addlinespace
    Sheep       & ear, eye, head, horn, leg, muzzle, neck, tail, torso \\
    \midrule
    \multicolumn{2}{c}{\textit{Base Categories (74 Parts)}} \\
    \midrule
    Aeroplane   & body, engine, stern, tail, wheel, wing \\
    \addlinespace
    Bicycle     & chainwheel, handlebar, headlight, saddle, wheel \\
    \addlinespace
    Bottle      & body, cap \\
    \addlinespace
    Bus         & back, door, front, headlight, license plate, mirror, roof, side, wheel, window \\
    \addlinespace
    Cat         & ear, eye, head, leg, neck, nose, paw, tail, torso \\
    \addlinespace
    Cow         & ear, eye, head, horn, leg, muzzle, neck, tail, torso \\
    \addlinespace
    Horse       & ear, eye, head, hoof, leg, muzzle, neck, tail, torso \\
    \addlinespace
    Person      & ear, eyebrow, eye, foot, hair, hand, head, leg, lower arm, mouth, neck, nose, torso, upper arm \\
    \addlinespace
    Pottedplant & plant, pot \\
    \addlinespace
    Train       & back, coach, front, head, headlight, roof, side \\
    \addlinespace
    Tvmonitor   & screen \\
\end{xltabular}

\textbf{ADE20K-Part-234.}
ADE20K-Part-234 is a customized iteration of the ADE20K dataset~\cite{ade20ksceneparsing}, optimized for OVPS challenges. This dataset provides a highly diverse set of object categories. Similar to Pascal-Part-116, the dataset is divided into base and novel classes. The designated novel object classes encompass ``bench'', ``bus'', ``fan'', ``desk'', ``stool'', ``truck'', ``van'', ``swivel chair'', ``oven'', ``ottoman'' and ``kitchen island''. The dataset contains 176 base and 58 novel categories. It introduces unique complexities for segmentation models due to the broad variety of classes and the prevalence of minute object parts that require high-precision mapping. Additional dataset details are provided in~\cref{tab:ade20k_part_234}. 

\begin{center}
\begin{xltabular}{\columnwidth}{@{}l >{\raggedright\arraybackslash}X@{}}
\caption{ADE20K-Part-234 base, novel categories and their associated parts.} \label{tab:ade20k_part_234} \\

\toprule
\textbf{Object} & \textbf{Parts} \\
\midrule
\endfirsthead

\toprule
\textbf{Object} & \textbf{Parts} \\
\midrule
\endhead

\midrule
\endfoot

\bottomrule
\endlastfoot

\multicolumn{2}{c}{\textit{Base Categories (176 Parts)}} \\
\cmidrule(r){1-2}
Airplane & door, fuselage, landing gear, propeller, stabilizer, turbine engine, wing \\
Armchair & apron, arm, back, back pillow, leg, seat, seat base, seat cushion \\
Bed & footboard, headboard, leg, side rail \\
Bookcase & door, drawer, front, side \\
Cabinet & door, drawer, front, shelf, side, skirt, top \\
Car & bumper, door, headlight, hood, license plate, logo, mirror, wheel, window, wiper \\
Chair & apron, arm, back, base, leg, seat, seat cushion, skirt, stretcher \\
Chandelier & arm, bulb, canopy, chain, cord, highlight, light source, shade \\
Chest of drawers & apron, door, drawer, front, leg \\
Clock & face, frame \\
Coffee table & leg, top \\
Computer & computer case, keyboard, monitor, mouse \\
Cooking stove & burner, button panel, door, drawer, oven, stove \\
Dishwasher & button panel, handle, skirt \\
Door & door frame, handle, knob, panel \\
Glass & base, bowl, opening, stem \\
Lamp & arm, base, canopy, column, cord, highlight, light source, shade, tube \\
Light & aperture, canopy, diffusor, highlight, light source, shade \\
Microwave & button panel, door, front, side, top, window \\
Minibike & license plate, mirror, seat, wheel \\
Person & arm, back, foot, gaze, hand, head, leg, neck, torso \\
Pool table & bed, leg, pocket \\
Refrigerator & button panel, door, drawer, side \\
Sconce & arm, backplate, highlight, light source, shade \\
Shelf & door, drawer, front, shelf \\
Sink & bowl, faucet, pedestal, tap, top \\
Sofa & arm, back, back pillow, leg, seat base, seat cushion, skirt \\
Table & apron, drawer, leg, shelf, top, wheel \\
Television receiver & base, buttons, frame, keys, screen, speaker \\
Toilet & bowl, cistern, lid \\
Traffic light & housing, pole \\
Wardrobe & door, drawer, front, leg, mirror, top \\
Washer & button panel, door, front, side \\
\midrule
\multicolumn{2}{c}{\textit{Novel Categories (58 Parts)}} \\
\cmidrule(r){1-2}
Bench & arm, back, leg, seat \\
Bus & bumper, door, headlight, license plate, logo, mirror, wheel, window, wiper \\
Desk & apron, door, drawer, leg, shelf, top \\
Fan & blade, canopy, tube \\
Kitchen island & door, drawer, front, side, top \\
Ottoman & back, leg, seat \\
Oven & button panel, door, drawer, top \\
Stool & leg, seat \\
Swivel chair & back, base, seat, wheel \\
Truck & bumper, door, headlight, license plate, logo, mirror, wheel, window \\
Van & bumper, door, headlight, license plate, logo, mirror, taillight, wheel, window, wiper \\
\end{xltabular}
\end{center}

\textbf{PartImageNet.}
Derived from ImageNet~\cite{imagenet}, PartImageNet~\cite{he2022partimagenet} encompasses roughly 24,000 images representing 158 distinct categories, complete with meticulous part-level annotations. Utilizing WordNet's hierarchical structure, these categories are organized into 11 broader superclasses. Because PartImageNet does not feature pre-established base and novel splits for cross-dataset testing, a zero-shot evaluation protocol is established by isolating 40 representative object categories which are further divided into 25 base and 15 novel object classes~\cite{choi2025fine}.

\begin{xltabular}{\textwidth}{@{}XXX@{}}
    \caption{PartImageNet base, novel object categories and their associated parts} \label{tab:partimagenet_custom_split} \\
    \toprule
    \textbf{Base Object Categories (25)} & \textbf{Novel Object Categories (15)} & \textbf{Part Classes} \\
    \midrule
    \endfirsthead

    \toprule
    \textbf{Base Object Categories (25)} & \textbf{Novel Object Categories (15)} & \textbf{Part Classes} \\
    \midrule
    \endhead

    \midrule
    \multicolumn{3}{r}{{Continued on next page\ldots}} \\
    \bottomrule
    \endfoot

    \bottomrule
    \endlastfoot

    Tiger, Giant panda, Leopard, Gazelle & Ice bear, Impala, Golden retriever & Head, Body, Foot, Tail \\
    \addlinespace
    Green mamba & Indian cobra & Head, Body \\
    \addlinespace
    Green lizard, Komodo dragon, Tree frog & Box turtle, American alligator & Head, Body, Foot, Tail \\
    \addlinespace
    Yawl, Pirate & Schooner & Body, Sail \\
    \addlinespace
    Barracouta, Goldfish, Killer whale & Tench & Head, Body, Fin, Tail \\
    \addlinespace
    Albatross, Goose & Bald eagle & Head, Body, Wing, Foot, Tail \\
    \addlinespace
    Garbage truck, Minibus, Ambulance & Jeep, School bus & Body, Tire, Side Mirror \\
    \addlinespace
    Mountain bike, Moped & Motor scooter & Body, Head, Seat, Tire \\
    \addlinespace
    Gorilla, Orangutan & Chimpanzee & Head, Body, Hand, Foot, Tail \\
    \addlinespace
    Beer bottle, Water bottle & Wine bottle & Mouth, Body \\
    \addlinespace
    Warplane & Airliner & Head, Body, Engine, Wing, Tail \\
\end{xltabular}

\textbf{PartImageNet OOD}
PartImageNet~\cite{he2022partimagenet} offers an alternative split designed for few-shot learning, which ensures non-overlapping classes across the training and validation sets. This few-shot split includes 109 base object classes in the training set and 19 novel object classes in the validation set. Detailed information about the base and novel classes along with their corresponding superclasses, is presented in~\cref{tab:partimagenet_ood}. 

\begin{xltabular}{\textwidth}{l X X}
\caption{PartImageNet (OOD) base, novel object categories and their associated parts.}
\label{tab:partimagenet_ood} \\

\toprule
\textbf{Superclass} & \textbf{Base Object Categories (109)} & \textbf{Novel Object Categories (19)} \\
\midrule
\endfirsthead

\toprule
\textbf{Superclass} & \textbf{Base Object Categories (109)} & \textbf{Novel Object Categories (19)} \\
\midrule
\endhead

\midrule
\multicolumn{3}{r}{{Continued on next page\ldots}} \\
\endfoot

\bottomrule
\endlastfoot

Bird & albatross, spoonbill, black stork, dowitcher, American egret, goose, ruddy turnstone, bee eater, kite & little blue heron, bald eagle \\
\addlinespace

Quadruped & impala, Egyptian cat, warthog, otter, Tibetan terrier, timber wolf, polecat, water buffalo, ox, redbone, English springer, tiger, American black bear, leopard, hartebeest, vizsla, Brittany spaniel, giant panda, Boston bull, ram, cairn, Arabian camel, fox squirrel, Eskimo dog, Irish water spaniel, Saluki, Walker hound, cheetah, gazelle, soft-coated wheaten terrier, bighorn, brown bear, chow, weasel & golden retriever, cougar, ice bear, mink, Saint Bernard \\
\addlinespace

Snake & night snake, boa constrictor, green mamba, thunder snake, green snake, hognose snake, sidewinder, horned viper, diamondback, rock python, garter snake, vine snake & Indian cobra \\
\addlinespace

Reptile & Gila monster, common newt, green lizard, bullfrog, American alligator, leatherback turtle, spotted salamander, box turtle, tailed frog, African chameleon, Komodo dragon, agama, frilled lizard, loggerhead & whiptail, alligator lizard \\
\addlinespace

Boat & yawl, pirate & trimaran \\
\addlinespace

Fish & goldfish, coho, tench, anemone fish, killer whale & barracouta, great white shark \\
\addlinespace

Car & garbage truck, minibus, ambulance, snowplow, golfcart, police van, minivan, convertible, limousine, recreational vehicle, go-kart, tractor, school bus, racer & beach wagon, cab \\
\addlinespace

Bicycle & motor scooter, tricycle, mountain bike & unicycle \\
\addlinespace

Biped & gorilla, gibbon, guenon, macaque, patas, howler monkey, chimpanzee, proboscis monkey, spider monkey, baboon, colobus, capuchin & siamang, marmoset \\
\addlinespace

Bottle & beer bottle, pop bottle, pill bottle & water bottle \\
\addlinespace

Aeroplane & airliner & \\

\end{xltabular}

\section{The Pascal-Part (org.) benchmark for fine-grained part instance segmentation}
  \label{sec:ovps_vs_pascalpart} 
\begin{figure}[h!] 
  \centering
  \includegraphics{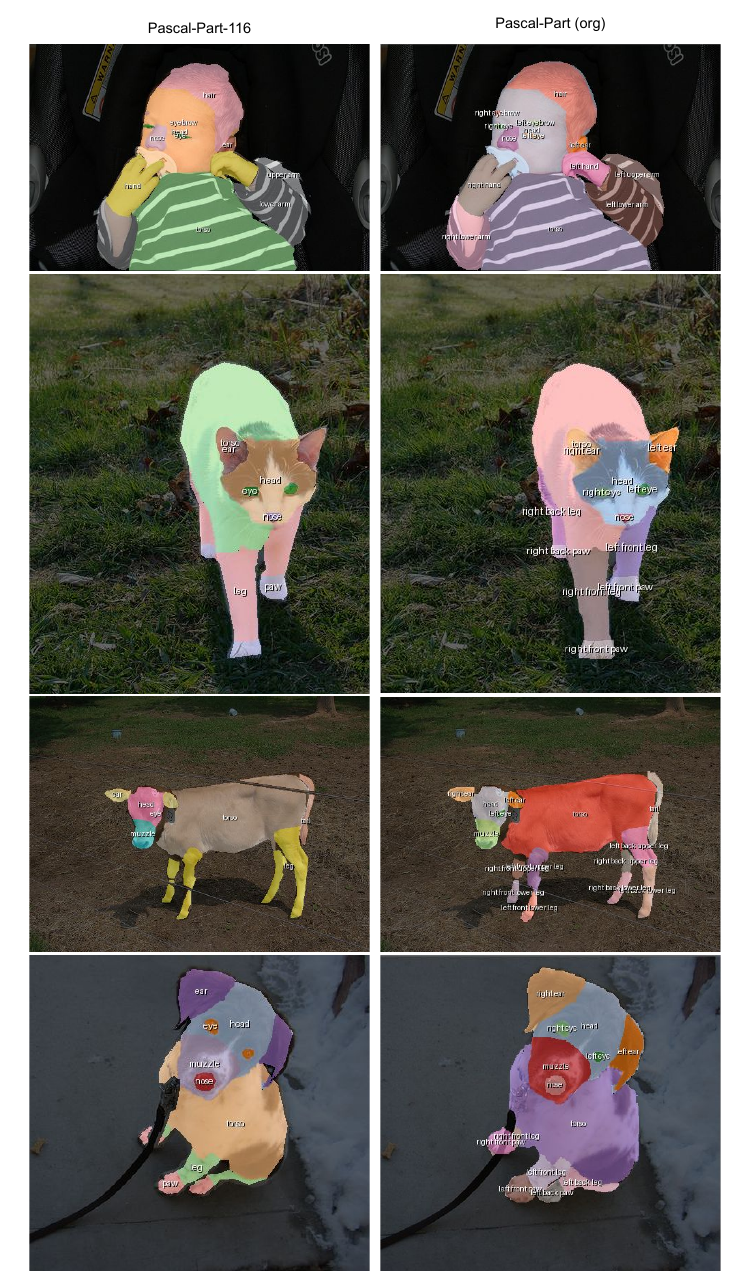} 
  \caption{\textbf{OVPS~\cite{ov-parts} vs. original Pascal-Part~\cite{pascalpart}.} The OVPS Pascal-Part-116~\cite{ov-parts} benchmark merges spatially distinct part instances from the original Pascal-Part~\cite{pascalpart}. For example, `left ear' and `right ear' are merged into a single `ear' instance.}
  \label{fig:ovps_vs_pascalpart} 
\end{figure}

The original Pascal-Part~\cite{pascalpart} consists of fine-grained part labels, with spatial identification of part instances (e.g., \underline{left lower} arm, \underline{right upper} arm). However, the Pascal-Part-116 (PPS-116)~\cite{ov-parts} split in the OVPS suite collapses these spatially distinct instances, e.g., `left ear' and `right ear', into a single `ear' instance (see~\cref{fig:ovps_vs_pascalpart}). This removes precision necessary for various real-world applications in robotics and T2I generation, reducing the problem from \textit{part instance segmentation} to \textit{part semantic segmentation}. Therefore, we additionally evaluate our approach on the original Pascal-Part~\cite{pascalpart} dataset in~\cref{subsec:pascalpart}, where our method significantly outperforms all baselines. 

\clearpage

\section{Part-Level IoU scores}
We present the part-level IoU scores for our model on the original Pascal-Part~\cite{pascalpart} benchmark in~\cref{tab:iou_parts_long}. 

\begin{xltabular}{\columnwidth}{@{}Xc c Xc c@{}}
    \caption{Part-Level IoU scores for our model on original Pascal-Part~\cite{pascalpart} benchmark.}\label{tab:iou_parts_long}\\
    \toprule
    \textbf{Part Name} & \textbf{IoU (\%)} & \textbf{Part Name} & \textbf{IoU (\%)} \\
    \midrule
    \endfirsthead

    \toprule
    \textbf{Part Name} & \textbf{IoU (\%)} & \textbf{Part Name} & \textbf{IoU (\%)} \\
    \midrule
    \endhead

    \midrule
    \multicolumn{4}{r}{{Continued on next page...}} \\
    \bottomrule
    \endfoot

    \bottomrule

    Bird beak & 38.31 & Cow right front lower leg & 65.10 \\
    Bird head & 66.78 & Cow right front upper leg & 53.92 \\
    Bird left eye & 44.32 & Cow right horn & 59.67 \\
    Bird left foot & 45.70 & Cow tail & 64.80 \\
    Bird left leg & 18.46 & Cow torso & 90.10 \\
    Bird left wing & 23.73 & Dog head & 66.45 \\
    Bird neck & 47.95 & Dog left back leg & 50.91 \\
    Bird right eye & 33.15 & Dog left back paw & 66.36 \\
    Bird right foot & 53.22 & Dog left ear & 67.84 \\
    Bird right leg & 23.58 & Dog left eye & 58.97 \\
    Bird right wing & 23.99 & Dog left front leg & 58.01 \\
    Bird tail & 52.86 & Dog left front paw & 67.58 \\
    Bird torso & 77.04 & Dog muzzle & 37.85 \\
    Bicycle back wheel & 16.96 & Dog neck & 51.89 \\
    Bicycle chain wheel & 72.35 & Dog nose & 14.89 \\
    Bicycle front wheel & 25.25 & Dog right back leg & 48.70 \\
    Bicycle handlebar & 90.34 & Dog right back paw & 56.96 \\
    Bicycle saddle & 28.48 & Dog right ear & 65.76 \\
    Bottle body & 98.37 & Dog right eye & 52.71 \\
    Bottle cap & 74.14 & Dog right front leg & 52.89 \\
    Bus back license plate & 0.00 & Dog right front paw & 70.37 \\
    Bus back side & 86.44 & Dog tail & 76.54 \\
    Bus door & 5.12 & Dog torso & 82.59 \\
    Bus front license plate & 0.05 & Motorbike back wheel & 9.30 \\
    Bus front side & 79.51 & Motorbike front wheel & 30.40 \\
    Bus headlight & 44.13 & Motorbike handlebar & 100.00 \\
    Bus left mirror & 47.61 & Motorbike headlight & 98.20 \\
    Bus left side & 78.63 & Pottedplant plant & 89.92 \\
    Bus right mirror & 22.92 & Aeroplane body & 85.10 \\
    Bus right side & 71.79 & Aeroplane engine & 32.73 \\
    Bus roof side & 23.51 & Aeroplane left wing & 58.83 \\
    Bus wheel & 67.91 & Aeroplane right wing & 55.07 \\
    Bus window & 81.83 & Aeroplane stern & 64.90 \\
    Car back license plate & 0.05 & Aeroplane tail & 0.01 \\
    Car back side & 72.68 & Aeroplane wheel & 84.53 \\
    Car door & 4.09 & Horse head & 76.62 \\
    Car front license plate & 0.01 & Horse left back hoof & 48.27 \\
    Car front side & 68.26 & Horse left back lower leg & 59.84 \\
    Car headlight & 40.81 & Horse left back upper leg & 56.14 \\
    Car left mirror & 40.84 & Horse left ear & 76.89 \\
    Car left side & 62.36 & Horse left eye & 42.65 \\
    Car right mirror & 53.83 & Horse left front hoof & 44.96 \\
    Car right side & 65.18 & Horse left front lower leg & 48.89 \\
    Car roof side & 25.36 & Horse left front upper leg & 52.91 \\
    Car wheel & 85.23 & Horse muzzle & 81.33 \\
    Car window & 73.71 & Horse neck & 74.18 \\
    Cat head & 84.15 & Horse right back hoof & 51.54 \\
    Cat left back leg & 34.10 & Horse right back lower leg & 63.90 \\
    Cat left back paw & 55.55 & Horse right back upper leg & 59.06 \\
    Cat left ear & 75.67 & Horse right ear & 72.05 \\
    Cat left eye & 57.15 & Horse right eye & 51.90 \\
    Cat left front leg & 47.68 & Horse right front hoof & 46.46 \\
    Cat left front paw & 55.64 & Horse right front lower leg & 51.18 \\
    Cat neck & 39.08 & Horse right front upper leg & 52.58 \\
    Cat nose & 44.14 & Horse tail & 78.49 \\
    Cat right back leg & 51.33 & Horse torso & 88.22 \\
    Cat right back paw & 63.84 & Person hair & 9.78 \\
    Cat right ear & 73.88 & Person head & 55.46 \\
    Cat right eye & 52.60 & Person left ear & 46.82 \\
    Cat right front leg & 46.43 & Person left eye & 34.89 \\
    Cat right front paw & 55.07 & Person left eyebrow & 25.47 \\
    Cat tail & 79.65 & Person left foot & 74.31 \\
    Cat torso & 83.85 & Person left hand & 71.61 \\
    Cow head & 70.53 & Person left lower arm & 69.44 \\
    Cow left back lower leg & 59.36 & Person left lower leg & 71.17 \\
    Cow left back upper leg & 38.47 & Person left upper arm & 77.45 \\
    Cow left ear & 70.87 & Person left upper leg & 72.54 \\
    Cow left eye & 54.49 & Person mouth & 50.51 \\
    Cow left front lower leg & 70.13 & Person neck & 61.79 \\
    Cow left front upper leg & 60.87 & Person nose & 56.53 \\
    Cow left horn & 74.17 & Person right ear & 57.70 \\
    Cow muzzle & 72.74 & Person right eye & 36.92 \\
    Cow neck & 48.46 & Person right eyebrow & 24.60 \\
    Cow right back lower leg & 66.76 & Person right foot & 69.60 \\
    Cow right back upper leg & 46.54 & Person right hand & 76.79 \\
    Cow right ear & 80.47 & Person right lower arm & 70.89 \\
    Cow right eye & 60.09 & Person right lower leg & 70.73 \\
    Sheep head & 61.07 & Person right upper arm & 75.25 \\
    Sheep left back lower leg & 63.56 & Person right upper leg & 71.44 \\
    Sheep left back upper leg & 39.47 & Person torso & 86.38 \\
    Sheep left ear & 58.56 & Pottedplant pot & 86.96 \\
    Sheep left eye & 32.48 & Train coach back side & 6.91 \\
    Sheep left front lower leg & 67.52 & Train coach front side & 0.10 \\
    Sheep left front upper leg & 59.81 & Train coach left side & 0.51 \\
    Sheep left horn & 42.43 & Train coach & 58.68 \\
    Sheep muzzle & 62.75 & Train coach right side & 27.52 \\
    Sheep neck & 51.07 & Train coach roof side & 33.68 \\
    Sheep right back lower leg & 64.42 & Train head & 5.09 \\
    Sheep right back upper leg & 49.52 & Train head back side & 19.04 \\
    Sheep right ear & 49.50 & Train head front side & 60.61 \\
    Sheep right eye & 26.24 & Train head left side & 53.95 \\
    Sheep right front lower leg & 64.84 & Train head right side & 62.64 \\
    Sheep right front upper leg & 61.21 & Train head roof side & 57.36 \\
    Sheep right horn & 46.81 & Train headlight & 1.55 \\
    Sheep tail & 51.58 & TV monitor screen & 100.00 \\
    Sheep torso & 90.67 & - & - \\
\end{xltabular}

\section{User study}
\label{section:supp_user_study}
The user study was conducted to evaluate attribute localization performance of our method against other baselines. The study was an A/B test where participants were shown two images, one generated by our method and the other by a baseline method. They were asked to identify which image was `better' w.r.t. our evaluation metric (attribute localization). 

\begin{figure}[ht!] 
  \centering
  \begin{minipage}[t]{0.48\linewidth}
    \centering
    \includegraphics[width=\linewidth]{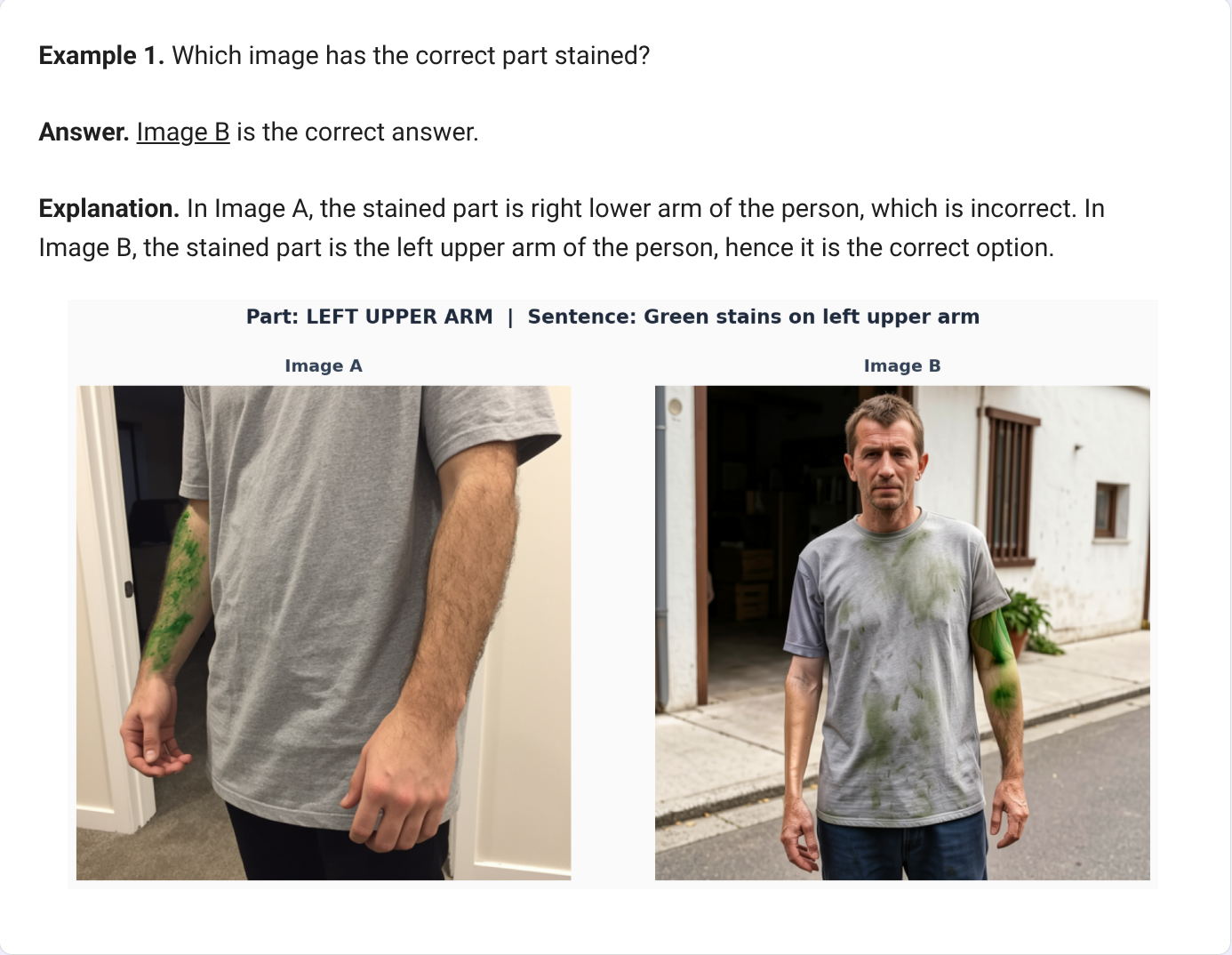}
  \end{minipage}\hfill
  \begin{minipage}[t]{0.48\linewidth}
    \centering
    \includegraphics[width=\linewidth]{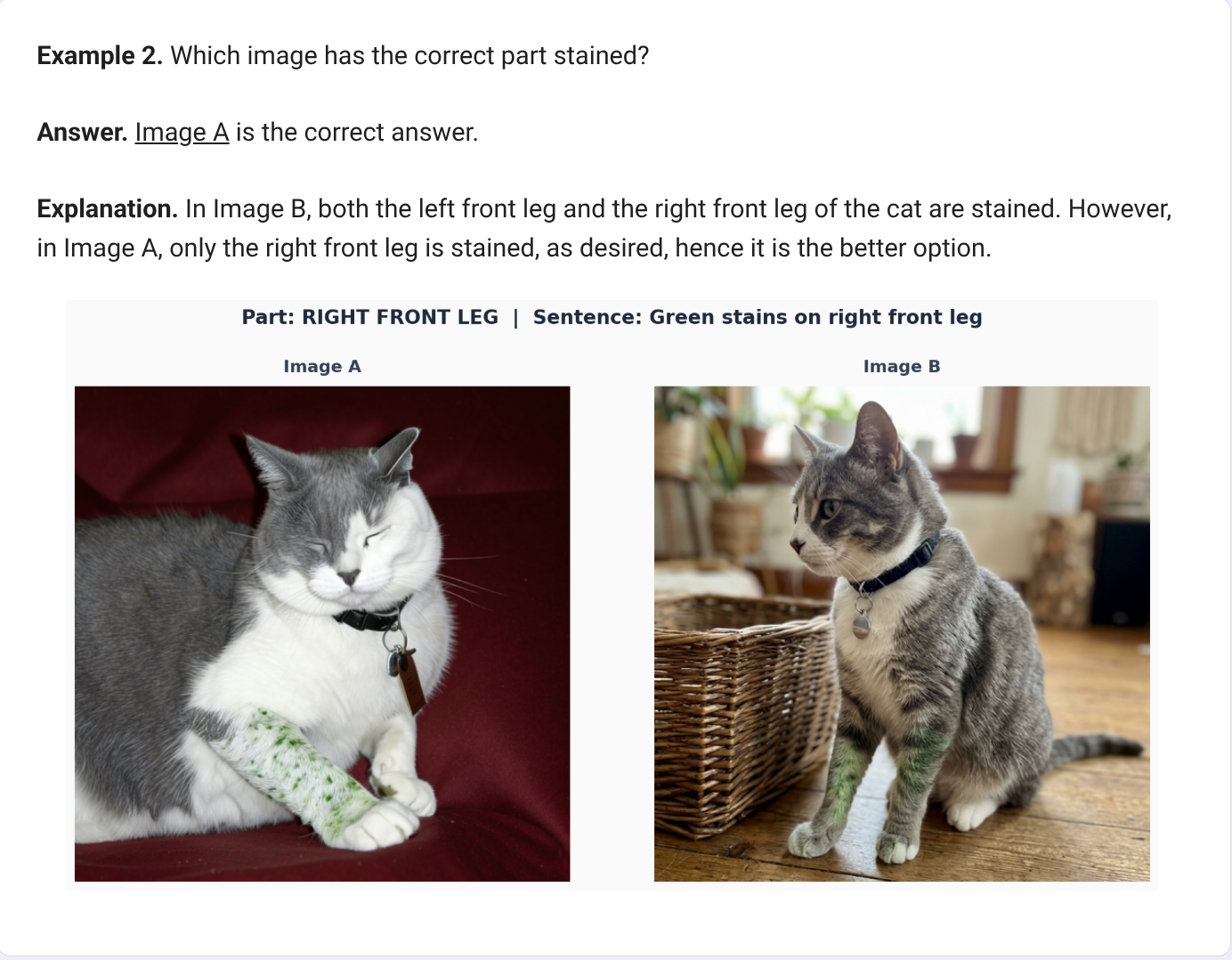}
  \end{minipage}
  \par\smallskip
  \begin{minipage}[t]{0.48\linewidth}
    \centering
    \includegraphics[width=\linewidth]{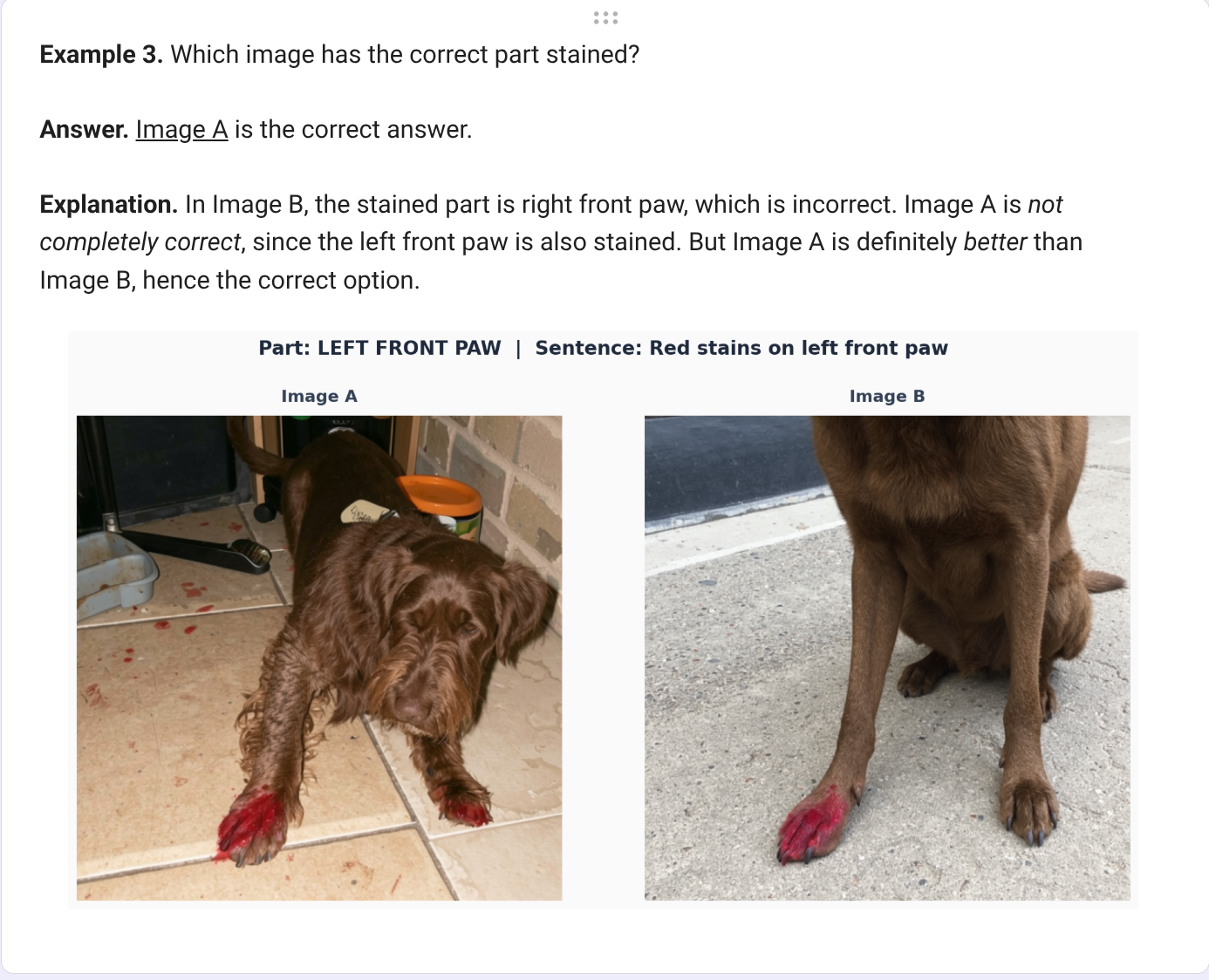}
  \end{minipage}\hfill
  \begin{minipage}[t]{0.48\linewidth}
    \centering
    \includegraphics[width=\linewidth]{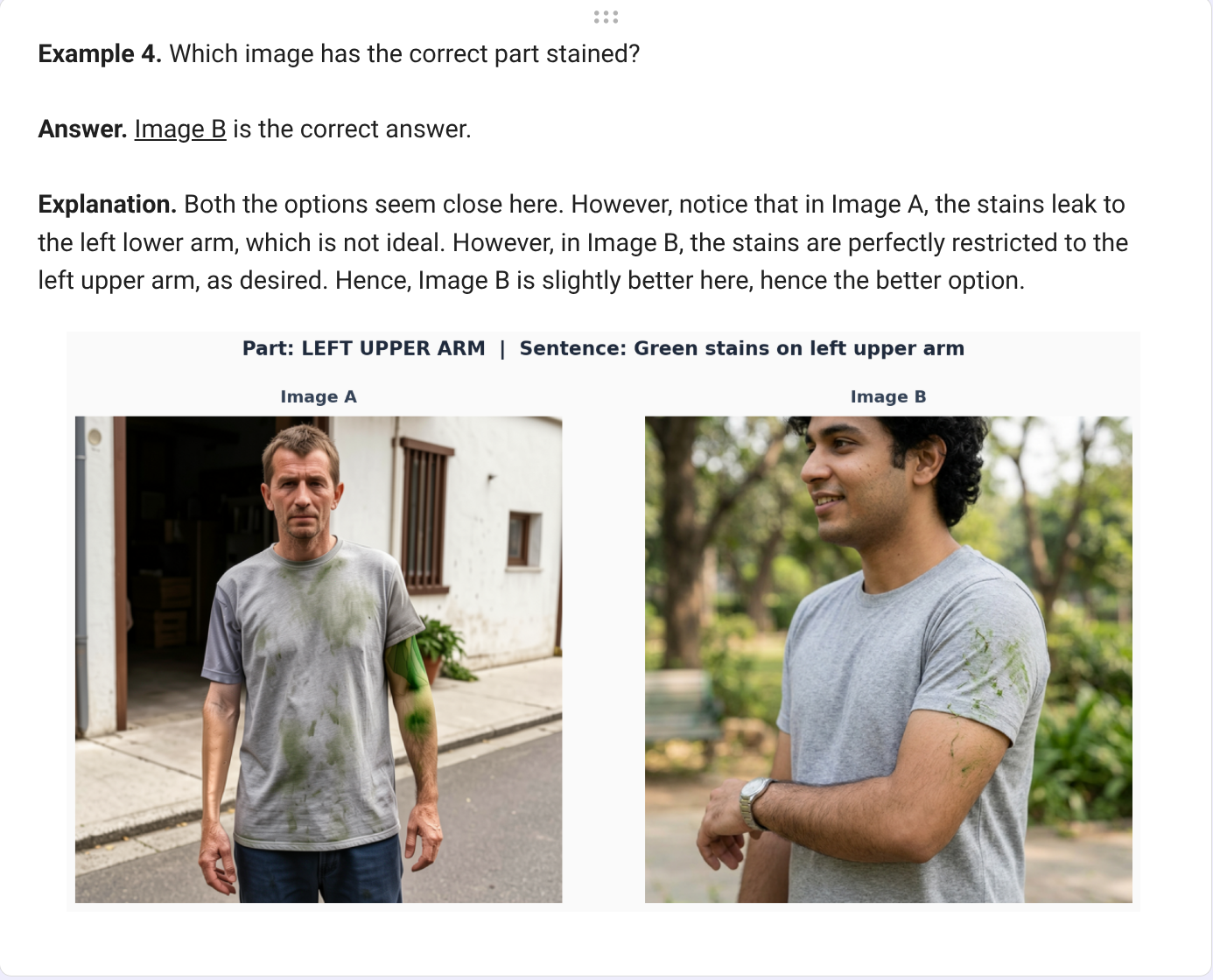}
  \end{minipage}
  \caption{Examples of questions shown to participants in our user study.}
  \label{fig:user_study_examples} 
\end{figure}

To explain the task to the participants, we provided four illustrative example questions, shown in \cref{fig:user_study_examples}.  

The study consisted of 51 questions, 17 questions per baseline method. We collected responses from 40 participants, resulting in a total of 2040 responses. The results of the user study are presented in \cref{fig:user_study}. 

\clearpage





\newpage
\section*{NeurIPS Paper Checklist}

\begin{enumerate}

\item {\bf Claims}
    \item[] Question: Do the main claims made in the abstract and introduction accurately reflect the paper's contributions and scope?
    \item[] Answer: \answerYes{} 
    \item[] Justification: In our abstract, we claim to improve text-aligned \textbf{part-level} understanding of a T2I model while preserving its prior. We validate this claim by showing that the improved part localization can be used for \textit{part-level segmentation} (\cref{subsec:experiments_ovps,subsec:pascalpart}). We also show that this strong localization enables precise part-level attribute binding, thus providing \textit{fine-grained instruction following} (~\cref{subsec:experiments_t2i}) in T2I generation.

\item {\bf Limitations}
    \item[] Question: Does the paper discuss the limitations of the work performed by the authors?
    \item[] Answer: \answerYes{} 
    \item[] Justification: The limitations are discussed in~\cref{sec:limitations} of the main paper.

\item {\bf Theory assumptions and proofs}
    \item[] Question: For each theoretical result, does the paper provide the full set of assumptions and a complete (and correct) proof?
    \item[] Answer: \answerNA{} 
    \item[] Justification: The paper does not include theoretical results.

\item {\bf Experimental result reproducibility}
    \item[] Question: Does the paper fully disclose all the information needed to reproduce the main experimental results of the paper to the extent that it affects the main claims and/or conclusions of the paper (regardless of whether the code and data are provided or not)?
    \item[] Answer: \answerYes{} 
    \item[] Justification: We describe the details of our method in ~\cref{sec:method}. We also describe implementation and training details in ~\cref{subsec:experiments_ovps} and~\cref{subsec:experiments_t2i}. Together these ensure reproducibility of our results.

\item {\bf Open access to data and code}
    \item[] Question: Does the paper provide open access to the data and code, with sufficient instructions to faithfully reproduce the main experimental results, as described in supplemental material?
    \item[] Answer: \answerNo{} 
    \item[] Justification: We will release both data and code post acceptance.

\item {\bf Experimental setting/details}
    \item[] Question: Does the paper specify all the training and test details (e.g., data splits, hyperparameters, how they were chosen, type of optimizer) necessary to understand the results?
    \item[] Answer: \answerYes{} 
    \item[] Justification: All details regarding data splits can be found in Appendix~\ref{sec:ovps_dataset_details}. All training and implementation details are mentioned in~\cref{subsec:experiments_ovps} and ~\cref{subsec:experiments_t2i}. 

\item {\bf Experiment statistical significance}
    \item[] Question: Does the paper report error bars suitably and correctly defined or other appropriate information about the statistical significance of the experiments?
    \item[] Answer: \answerNo{} 
    \item[] Justification: Error bars are not reported because it would be too computationally expensive. 

\item {\bf Experiments compute resources}
    \item[] Question: For each experiment, does the paper provide sufficient information on the computer resources (type of compute workers, memory, time of execution) needed to reproduce the experiments?
    \item[] Answer: \answerYes{} 
    \item[] Justification: Our compute requirements are mentioned in Appendix ~\ref{sec:compute_resources}
    
\item {\bf Code of ethics}
    \item[] Question: Does the research conducted in the paper conform, in every respect, with the NeurIPS Code of Ethics \url{https://neurips.cc/public/EthicsGuidelines}?
    \item[] Answer: \answerYes{} 
    \item[] Justification: The paper conforms to the NeurIPS Code of Ethics.

\item {\bf Broader impacts}
    \item[] Question: Does the paper discuss both potential positive societal impacts and negative societal impacts of the work performed?
    \item[] Answer: \answerNo{} 
    \item[] Justification: Our work focuses on the fundamental technical aspects of part-level understanding in Text-to-Image (T2I) models. While we acknowledge the broad societal implications of T2I technology, this research is a controlled diagnostic study that does not introduce new risks/impact. Consequently, a standalone discussion of societal impact was omitted in favor of focusing on the technical analysis. 
    
\item {\bf Safeguards}
    \item[] Question: Does the paper describe safeguards that have been put in place for responsible release of data or models that have a high risk for misuse (e.g., pre-trained language models, image generators, or scraped datasets)?
    \item[] Answer: \answerYes{} 
    \item[] Justification: Our model inherits safeguards from the base model~\cite{flux-2-2025} upon which our model has been trained.

\item {\bf Licenses for existing assets}
    \item[] Question: Are the creators or original owners of assets (e.g., code, data, models), used in the paper, properly credited and are the license and terms of use explicitly mentioned and properly respected?
    \item[] Answer: \answerYes{} 
    \item[] Justification: The models and data used in the paper are properly credited, the license and terms of use are correctly respected.

\item {\bf New assets}
    \item[] Question: Are new assets introduced in the paper well documented and is the documentation provided alongside the assets?
    \item[] Answer: \answerYes{} 
    \item[] Justification: Documentation for the assets is already provided throughout the paper's content. The assets will be made publicly available post acceptance.

\item {\bf Crowdsourcing and research with human subjects}
    \item[] Question: For crowdsourcing experiments and research with human subjects, does the paper include the full text of instructions given to participants and screenshots, if applicable, as well as details about compensation (if any)? 
    \item[] Answer: \answerYes{} 
    \item[] Justification: We conducted a user study to evaluate generation accuracy. Participation was entirely voluntary, no compensation was given. A screenshot of the user study has been added in Appendix ~\ref{section:supp_user_study}

\item {\bf Institutional review board (IRB) approvals or equivalent for research with human subjects}
    \item[] Question: Does the paper describe potential risks incurred by study participants, whether such risks were disclosed to the subjects, and whether Institutional Review Board (IRB) approvals (or an equivalent approval/review based on the requirements of your country or institution) were obtained?
    \item[] Answer: \answerYes{} 
    \item[] Justification: The paper involves a user study for evaluating generation accuracy; this does not involve any risks to the participants. We have received the necessary IRB approvals for this.  

\item {\bf Declaration of LLM usage}
    \item[] Question: Does the paper describe the usage of LLMs if it is an important, original, or non-standard component of the core methods in this research? Note that if the LLM is used only for writing, editing, or formatting purposes and does \emph{not} impact the core methodology, scientific rigor, or originality of the research, declaration is not required.
    \item[] Answer: \answerYes{} 
    \item[] Justification: A dataset preparation stage requires usage of VLMs for captioning images, see Appendix~\ref{sec:vlm_part_captioning}. We ensure that our process adheres to the NeurIPS code.  

\end{enumerate}

\end{document}